# Gaussian Process Optimization in the Bandit Setting: No Regret and Experimental Design

Niranjan Srinivas
California Institute of Technology
niranjan@caltech.edu

Andreas Krause
California Institute of Technology
krausea@caltech.edu

Sham M. Kakade
University of Pennsylvania
skakade@wharton.upenn.edu

Matthias Seeger
Saarland University
mseeger@mmci.uni-saarland.de

## Abstract

Many applications require optimizing an unknown, noisy function that is expensive to evaluate. We formalize this task as a multi-armed bandit problem, where the payoff function is either sampled from a Gaussian process (GP) or has low RKHS norm. We resolve the important open problem of deriving regret bounds for this setting, which imply novel convergence rates for GP optimization. We analyze GP-UCB, an intuitive upper-confidence based algorithm, and bound its cumulative regret in terms of maximal information gain, establishing a novel connection between GP optimization and experimental design. Moreover, by bounding the latter in terms of operator spectra, we obtain explicit sublinear regret bounds for many commonly used covariance functions. In some important cases, our bounds have surprisingly weak dependence on the dimensionality. In our experiments on real sensor data, GP-UCB compares favorably with other heuristical GP optimization approaches.

## 1. Introduction

In most stochastic optimization settings, evaluating the unknown function is expensive, and sampling is to be minimized. Examples include choosing advertisements in sponsored search to maximize profit in a click-through model (Pandey & Olston, 2007) or learning optimal control strategies for robots (Lizotte et al., 2007). Predominant approaches to this problem include the multi-armed bandit paradigm (Robbins, 1952), where the goal is to maximize cumulative reward by optimally balancing exploration and exploitation, and experimental design (Chaloner & Verdinelli, 1995), where the function is to be explored globally with as few evaluations as possible, for example by maximizing information gain. The challenge in both approaches is twofold: we have to estimate an unknown function $f$ from noisy samples, and we must optimize our estimate over some high-dimensional input space. For the former, much progress has been made in machine learning through kernel methods and Gaussian process (GP) models (Rasmussen & Williams, 2006), where smoothness assumptions about $f$ are encoded through the choice of kernel in a flexible nonparametric fashion. Beyond Euclidean spaces, kernels can be defined on diverse domains such as spaces of graphs, sets, or lists.

---

[1]This is the longer version of our paper in ICML 2010; see Srinivas et al. (2010)

We are concerned with GP optimization in the multi-armed bandit setting, where $f$ is sampled from a GP distribution or has low "complexity" measured in terms of its RKHS norm under some kernel. We provide the first sublinear regret bounds in this nonparametric setting, which imply convergence rates for GP optimization. In particular, we analyze the Gaussian Process Upper Confidence Bound (GP-UCB) algorithm, a simple and intuitive Bayesian method (Auer et al., 2002; Auer, 2002; Dani et al., 2008). While objectives are different in the multi-armed bandit and experimental design paradigm, our results draw a close technical connection between them: our regret bounds come in terms of an *information gain* quantity, measuring how fast $f$ can be learned in an information theoretic sense. The submodularity of this function allows us to prove sharp regret bounds for particular covariance functions, which we demonstrate for commonly used Squared Exponential and Matérn kernels.

**Related Work.** Our work generalizes stochastic *linear* optimization in a bandit setting, where the unknown function comes from a finite-dimensional linear space. GPs are nonlinear random functions, which can be represented in an infinite-dimensional linear space. For the standard linear setting, Dani et al. (2008) provide a near-complete characterization (also see Auer 2002; Dani et al. 2007; Abernethy et al. 2008; Rusmevichientong & Tsitsiklis 2008), explicitly dependent on the dimensionality. In the GP setting, the challenge is to characterize complexity in a different manner, through properties of the kernel function. Our technical contributions are twofold: first, we show how to analyze the nonlinear setting by focusing on the concept of information gain, and second, we explicitly bound this information gain measure using the concept of submodularity (Nemhauser et al., 1978) and knowledge about kernel operator spectra.

Kleinberg et al. (2008) provide regret bounds under weaker and less configurable assumptions (only Lipschitz-continuity w.r.t. a metric is assumed; Bubeck et al. 2008 consider arbitrary topological spaces), which however degrade rapidly with the dimensionality of the problem ($\Omega(T^{\frac{d+1}{d+2}})$). In practice, linearity w.r.t. a fixed basis is often too stringent an assumption, while Lipschitz-continuity can be too coarse-grained, leading to poor rate bounds. Adopting GP assumptions, we can model levels of smoothness in a fine-grained way. For example, our rates for the frequently used Squared Exponential kernel, enforcing a high degree of smoothness, have weak dependence on the dimensionality: $\mathcal{O}(\sqrt{T(\log T)^{d+1}})$ (see Fig. 1).

There is a large literature on GP (response surface) optimization. Several heuristics for trading off exploration and exploitation in GP optimization have been proposed (such as Expected Improvement, Mockus et al. 1978, and Most Probable Improvement, Mockus 1989) and successfully applied in practice (*c.f.*, Lizotte et al. 2007). Brochu et al. (2009) provide a comprehensive review of and motivation for Bayesian optimization using GPs. The Efficient Global Optimization (EGO) algorithm for optimizing expensive black-box functions is proposed by Jones et al. (1998) and extended to GPs by Huang et al. (2006). Little is known about theoretical performance of GP optimization. While convergence of EGO is established by Vazquez & Bect (2007), convergence rates have remained elusive. Grünewälder et al. (2010) consider the pure exploration problem for GPs, where the goal is to find the optimal decision over $T$ rounds, rather than maximize cumulative reward (with no exploration/exploitation dilemma). They provide sharp bounds for this exploration problem. Note that this methodology would not lead to bounds for minimizing the cumulative regret. Our cumulative regret bounds translate to the first performance guarantees (rates) for GP optimization.

**Summary.** Our main contributions are:

- We analyze GP-UCB, an intuitive algorithm for GP optimization, when the function is either sampled from a known GP, or has low RKHS norm.
- We bound the cumulative regret for GP-UCB in terms of the information gain due to sampling, establishing a novel connection between experimental design and GP optimization.
- By bounding the information gain for popular classes of kernels, we establish sublinear regret bounds for GP optimization for the first time. Our bounds depend on kernel choice and parameters in a fine-grained fashion.
- We evaluate GP-UCB on sensor network data, demonstrating that it compares favorably to existing algorithms for GP optimization.

| *Kernel* | Linear | RBF | Matérn |
|---|---|---|---|
| **Regret** $R_T$ | $d\sqrt{T}$ | $\sqrt{T(\log T)^{d+1}}$ | $T^{\frac{\nu+d(d+1)}{2\nu+d(d+1)}}$ |

*Figure 1.* Our regret bounds (up to polylog factors) for linear, radial basis, and Matérn kernels — $d$ is the dimension, $T$ is the time horizon, and $\nu$ is a Matérn parameter.

## 2. Problem Statement and Background

Consider the problem of sequentially optimizing an unknown reward function $f : D \to \mathbb{R}$: in each round $t$, we choose a point $\boldsymbol{x}_t \in D$ and get to see the function value there, perturbed by noise: $y_t = f(\boldsymbol{x}_t) + \epsilon_t$. Our goal is to maximize the sum of rewards $\sum_{t=1}^T f(\boldsymbol{x}_t)$, thus to perform essentially as well as $\boldsymbol{x}^* = \operatorname{argmax}_{\boldsymbol{x} \in D} f(\boldsymbol{x})$ (as rapidly as possible). For example, we might want to find locations of highest temperature in a building by sequentially activating sensors in a spatial network and regressing on their measurements. $D$ consists of all sensor locations, $f(\boldsymbol{x})$ is the temperature at $\boldsymbol{x}$, and sensor accuracy is quantified by the noise variance. Each activation draws battery power, so we want to sample from as few sensors as possible.

**Regret.** A natural performance metric in this context is cumulative regret, the loss in reward due to not knowing $f$'s maximum points beforehand. Suppose the unknown function is $f$, its maximum point[1] $\boldsymbol{x}^* = \operatorname{argmax}_{\boldsymbol{x} \in D} f(\boldsymbol{x})$. For our choice $\boldsymbol{x}_t$ in round $t$, we incur instantaneous regret $r_t = f(\boldsymbol{x}^*) - f(\boldsymbol{x}_t)$. The *cumulative regret* $R_T$ after $T$ rounds is the sum of instantaneous regrets: $R_T = \sum_{t=1}^T r_t$. A desirable asymptotic property of an algorithm is to be *no-regret*: $\lim_{T \to \infty} R_T/T = 0$. Note that neither $r_t$ nor $R_T$ are ever revealed to the algorithm. Bounds on the average regret $R_T/T$ translate to convergence rates for GP optimization: the maximum $\max_{t \leq T} f(\boldsymbol{x}_t)$ in the first $T$ rounds is no further from $f(\boldsymbol{x}^*)$ than the average.

---

[1] $\boldsymbol{x}^*$ need not be unique; only $f(\boldsymbol{x}^*)$ occurs in the regret.

## 2.1. Gaussian Processes and RKHS's

**Gaussian Processes.** Some assumptions on $f$ are required to guarantee no-regret. While rigid parametric assumptions such as linearity may not hold in practice, a certain degree of smoothness is often warranted. In our sensor network, temperature readings at closeby locations are highly correlated (see Figure 2(a)). We can enforce implicit properties like smoothness without relying on any parametric assumptions, modeling $f$ as a sample from a *Gaussian process* (GP): a collection of dependent random variables, one for each $\boldsymbol{x} \in D$, every finite subset of which is multivariate Gaussian distributed in an overall consistent way (Rasmussen & Williams, 2006). A $GP(\mu(\boldsymbol{x}), k(\boldsymbol{x}, \boldsymbol{x}'))$ is specified by its mean function $\mu(\boldsymbol{x}) = \mathbb{E}[f(\boldsymbol{x})]$ and covariance (or kernel) function $k(\boldsymbol{x}, \boldsymbol{x}') = \mathbb{E}[(f(x) - \mu(\boldsymbol{x}))(f(x') - \mu(\boldsymbol{x}'))]$. For GPs not conditioned on data, we assume[2] that $\mu \equiv 0$. Moreover, we restrict $k(\boldsymbol{x}, \boldsymbol{x}) \leq 1$, $\boldsymbol{x} \in D$, i.e., we assume bounded variance. By fixing the correlation behavior, the covariance function $k$ encodes smoothness properties of sample functions $f$ drawn from the GP. A range of commonly used kernel functions is given in Section 5.2.

In this work, GPs play multiple roles. First, some of our results hold when the unknown target function is a sample from a known GP distribution $GP(0, k(\boldsymbol{x}, \boldsymbol{x}'))$. Second, the Bayesian algorithm we analyze generally uses $GP(0, k(\boldsymbol{x}, \boldsymbol{x}'))$ as prior distribution over $f$. A major advantage of working with GPs is the existence of simple analytic formulae for mean and covariance of the posterior distribution, which allows easy implementation of algorithms. For a noisy sample $\boldsymbol{y}_T = [y_1 \ldots y_T]^T$ at points $A_T = \{\boldsymbol{x}_1, \ldots, \boldsymbol{x}_T\}$, $y_t = f(\boldsymbol{x}_t) + \epsilon_t$ with $\epsilon_t \sim N(0, \sigma^2)$ i.i.d. Gaussian noise, the posterior over $f$ is a GP distribution again, with mean $\mu_T(\boldsymbol{x})$, covariance $k_T(\boldsymbol{x}, \boldsymbol{x}')$ and variance $\sigma_T^2(\boldsymbol{x})$:

$$\mu_T(\boldsymbol{x}) = \boldsymbol{k}_T(\boldsymbol{x})^T (\boldsymbol{K}_T + \sigma^2 \boldsymbol{I})^{-1} \boldsymbol{y}_T, \quad (1)$$

$$k_T(\boldsymbol{x}, \boldsymbol{x}') = k(\boldsymbol{x}, \boldsymbol{x}') - \boldsymbol{k}_T(\boldsymbol{x})^T (\boldsymbol{K}_T + \sigma^2 \boldsymbol{I})^{-1} \boldsymbol{k}_T(\boldsymbol{x}'),$$

$$\sigma_T^2(\boldsymbol{x}) = k_T(\boldsymbol{x}, \boldsymbol{x}), \quad (2)$$

where $\boldsymbol{k}_T(\boldsymbol{x}) = [k(\boldsymbol{x}_1, \boldsymbol{x}) \ldots k(\boldsymbol{x}_T, \boldsymbol{x})]^T$ and $\boldsymbol{K}_T$ is the positive definite kernel matrix $[k(\boldsymbol{x}, \boldsymbol{x}')]_{\boldsymbol{x}, \boldsymbol{x}' \in A_T}$.

**RKHS.** Instead of the Bayes case, where $f$ is sampled from a GP prior, we also consider the more agnostic case where $f$ has low "complexity" as measured under an RKHS norm (and distribution free assumptions on the noise process). The notion of *reproducing kernel Hilbert spaces* (RKHS, Wahba 1990) is intimately related to GPs and their covariance functions $k(\boldsymbol{x}, \boldsymbol{x}')$. The RKHS $\mathcal{H}_k(D)$ is a complete subspace of $L_2(D)$ of nicely behaved functions, with an inner product $\langle \cdot, \cdot \rangle_k$ obeying the reproducing property: $\langle f, k(\boldsymbol{x}, \cdot) \rangle_k = f(\boldsymbol{x})$ for all $f \in \mathcal{H}_k(D)$. It is literally constructed by completing the set of mean functions $\mu_T$ for all possible $T$, $\{\boldsymbol{x}_t\}$, and $\boldsymbol{y}_T$. The induced RKHS norm $\|f\|_k = \sqrt{\langle f, f \rangle_k}$ measures smoothness of $f$ w.r.t. $k$: in much the same way as $k_1$ would generate smoother samples than $k_2$ as GP covariance functions, $\|\cdot\|_{k_1}$ assigns larger penalties than $\|\cdot\|_{k_2}$. $\langle \cdot, \cdot \rangle_k$ can be extended to all of $L_2(D)$, in which case $\|f\|_k < \infty$ iff $f \in \mathcal{H}_k(D)$. For most kernels discussed in Section 5.2, members of $\mathcal{H}_k(D)$ can uniformly approximate any continuous function on any compact subset of $D$.

## 2.2. Information Gain & Experimental Design

One approach to maximizing $f$ is to first choose points $\boldsymbol{x}_t$ so as to estimate the function globally well, then play the maximum point of our estimate. How can we learn about $f$ as rapidly as possible? This question comes down to Bayesian Experimental Design (henceforth "ED"; see Chaloner & Verdinelli 1995), where the informativeness of a set of sampling points $A \subset D$ about $f$ is measured by the *information gain* (c.f., Cover & Thomas 1991), which is the mutual information between $f$ and observations $\boldsymbol{y}_A = \boldsymbol{f}_A + \epsilon_A$ at these points:

$$I(\boldsymbol{y}_A; f) = H(\boldsymbol{y}_A) - H(\boldsymbol{y}_A | f), \quad (3)$$

quantifying the reduction in uncertainty about $f$ from revealing $\boldsymbol{y}_A$. Here, $\boldsymbol{f}_A = [f(\boldsymbol{x})]_{\boldsymbol{x} \in A}$ and $\varepsilon_A \sim N(\boldsymbol{0}, \sigma^2 \boldsymbol{I})$. For a Gaussian, $H(N(\boldsymbol{\mu}, \boldsymbol{\Sigma})) = \frac{1}{2} \log |2\pi e \boldsymbol{\Sigma}|$, so that in our setting $I(\boldsymbol{y}_A; f) = I(\boldsymbol{y}_A; \boldsymbol{f}_A) = \frac{1}{2} \log |\boldsymbol{I} + \sigma^{-2} \boldsymbol{K}_A|$, where $\boldsymbol{K}_A = [k(\boldsymbol{x}, \boldsymbol{x}')]_{\boldsymbol{x}, \boldsymbol{x}' \in A}$. While finding the information gain maximizer among $A \subset D$, $|A| \leq T$ is NP-hard (Ko et al., 1995), it can be approximated by an efficient greedy algorithm. If $F(A) = I(\boldsymbol{y}_A; f)$, this algorithm picks $\boldsymbol{x}_t = \operatorname{argmax}_{\boldsymbol{x} \in D} F(A_{t-1} \cup \{\boldsymbol{x}\})$ in round $t$, which can be shown to be equivalent to

$$\boldsymbol{x}_t = \underset{\boldsymbol{x} \in D}{\operatorname{argmax}} \, \sigma_{t-1}(\boldsymbol{x}), \quad (4)$$

where $A_{t-1} = \{\boldsymbol{x}_1, \ldots, \boldsymbol{x}_{t-1}\}$. Importantly, this simple algorithm is guaranteed to find a near-optimal solution: for the set $A_T$ obtained after $T$ rounds, we have that

$$F(A_T) \geq (1 - 1/e) \max_{|A| \leq T} F(A), \quad (5)$$

at least a constant fraction of the optimal information gain value. This is because $F(A)$ satisfies a diminishing returns property called *submodularity* (Krause & Guestrin, 2005), and the greedy approximation guarantee (5) holds for any submodular function (Nemhauser et al., 1978).

While sequentially optimizing Eq. 4 is a provably good way to *explore* $f$ globally, it is not well suited for func-tion optimization. For the latter, we only need to identify points $\boldsymbol{x}$ where $f(\boldsymbol{x})$ is large, in order to concentrate sampling there as rapidly as possible, thus *exploit* our knowledge about maxima. In fact, the ED rule (4) does not even depend on observations $y_t$ obtained along the way. Nevertheless, the maximum information gain after $T$ rounds will play a prominent role in our regret bounds, forging an important connection between GP optimization and experimental design.

---
[2]This is w.l.o.g. (Rasmussen & Williams, 2006).

## 3. GP-UCB Algorithm

For sequential optimization, the ED rule (4) can be wasteful: it aims at decreasing uncertainty globally, not just where maxima might be. Another idea is to pick points as $\boldsymbol{x}_t = \operatorname{argmax}_{\boldsymbol{x} \in D} \mu_{t-1}(\boldsymbol{x})$, maximizing the expected reward based on the posterior so far. However, this rule is too greedy too soon and tends to get stuck in shallow local optima. A combined strategy is to choose

$$\boldsymbol{x}_t = \operatorname*{argmax}_{\boldsymbol{x} \in D} \mu_{t-1}(\boldsymbol{x}) + \beta_t^{1/2} \sigma_{t-1}(\boldsymbol{x}), \qquad (6)$$

where $\beta_t$ are appropriate constants. This latter objective prefers both points $\boldsymbol{x}$ where $f$ is uncertain (large $\sigma_{t-1}(\cdot)$) and such where we expect to achieve high rewards (large $\mu_{t-1}(\cdot)$): it implicitly negotiates the exploration–exploitation tradeoff. A natural interpretation of this sampling rule is that it greedily selects points $\boldsymbol{x}$ such that $f(\boldsymbol{x})$ should be a reasonable upper bound on $f(\boldsymbol{x}^*)$, since the argument in (6) is an upper quantile of the marginal posterior $P(f(\boldsymbol{x})|\boldsymbol{y}_{t-1})$. We call this choice the *Gaussian process upper confidence bound* rule (GP-UCB), where $\beta_t$ is specified depending on the context (see Section 4). Pseudocode for the GP-UCB algorithm is provided in Algorithm 1. Figure 2 illustrates two subsequent iterations, where GP-UCB both explores (Figure 2(b)) by sampling an input $\boldsymbol{x}$ with large $\sigma_{t-1}^2(\boldsymbol{x})$ and exploits (Figure 2(c)) by sampling $\boldsymbol{x}$ with large $\mu_{t-1}(\boldsymbol{x})$.

The GP-UCB selection rule Eq. 6 is motivated by the UCB algorithm for the classical multi-armed bandit problem (Auer et al., 2002; Kocsis & Szepesvári, 2006). Among competing criteria for GP optimization (see Section 1), a variant of the GP-UCB rule has been demonstrated to be effective for this application (Dorard et al., 2009). To our knowledge, strong theoretical results of the kind provided for GP-UCB in this paper have not been given for any of these search heuristics. In Section 6, we show that in practice GP-UCB compares favorably with these alternatives.

If $D$ is infinite, finding $\boldsymbol{x}_t$ in (6) may be hard: the upper confidence index is multimodal in general. However, global search heuristics are very effective in practice (Brochu et al., 2009). It is generally assumed

---

**Algorithm 1** The GP-UCB algorithm.

**Input:** Input space $D$; GP Prior $\mu_0 = 0$, $\sigma_0$, $k$
**for** $t = 1, 2, \ldots$ **do**
  Choose $\boldsymbol{x}_t = \operatorname*{argmax}_{\boldsymbol{x} \in D} \mu_{t-1}(\boldsymbol{x}) + \sqrt{\beta_t} \sigma_{t-1}(\boldsymbol{x})$
  Sample $y_t = f(\boldsymbol{x}_t) + \epsilon_t$
  Perform Bayesian update to obtain $\mu_t$ and $\sigma_t$
**end for**

---

that evaluating $f$ is more costly than maximizing the UCB index.

UCB algorithms (and GP optimization techniques in general) have been applied to a large number of problems in practice (Kocsis & Szepesvári, 2006; Pandey & Olston, 2007; Lizotte et al., 2007). Their performance is well characterized in both the finite arm setting and the linear optimization setting, but no convergence rates for GP optimization are known.

## 4. Regret Bounds

We now establish cumulative regret bounds for GP optimization, treating a number of different settings: $f \sim \operatorname{GP}(0, k(\boldsymbol{x}, \boldsymbol{x}'))$ for finite $D$, $f \sim \operatorname{GP}(0, k(\boldsymbol{x}, \boldsymbol{x}'))$ for general compact $D$, and the agnostic case of arbitrary $f$ with bounded RKHS norm.

GP optimization generalizes stochastic linear optimization, where a function $f$ from a finite-dimensional linear space is optimized over. For the linear case, Dani et al. (2008) provide regret bounds that explicitly depend on the dimensionality[3] $d$. GPs can be seen as random functions in some infinite-dimensional linear space, so their results do not apply in this case. This problem is circumvented in our regret bounds. The quantity governing them is the *maximum information gain* $\gamma_T$ after $T$ rounds, defined as:

$$\gamma_T := \max_{A \subset D : |A| = T} \operatorname{I}(\boldsymbol{y}_A; \boldsymbol{f}_A), \qquad (7)$$

where $\operatorname{I}(\boldsymbol{y}_A; \boldsymbol{f}_A) = \operatorname{I}(\boldsymbol{y}_A; f)$ is defined in (3). Recall that $\operatorname{I}(\boldsymbol{y}_A; \boldsymbol{f}_A) = \frac{1}{2} \log |\boldsymbol{I} + \sigma^{-2} \boldsymbol{K}_A|$, where $\boldsymbol{K}_A = [k(\boldsymbol{x}, \boldsymbol{x}')]_{\boldsymbol{x}, \boldsymbol{x}' \in A}$ is the covariance matrix of $\boldsymbol{f}_A = [f(\boldsymbol{x})]_{\boldsymbol{x} \in A}$ associated with the samples $A$. Our regret bounds are of the form $\mathcal{O}^*(\sqrt{T \beta_T \gamma_T})$, where $\beta_T$ is the confidence parameter in Algorithm 1, while the bounds of Dani et al. (2008) are of the form $\mathcal{O}^*(\sqrt{T \beta_T} d)$ ($d$ the dimensionality of the linear function space). Here and below, the $\mathcal{O}^*$ notation is a variant of $\mathcal{O}$, where log factors are suppressed. While our proofs – all provided in the Appendix – use techniques similar to those of Dani et al. (2008), we face a number of additional significant technical challenges. Besides avoiding the finite-dimensional analysis, we must handle confidence issues, which are more delicate for nonlinear random functions.

---

[3] In general, $d$ is the dimensionality of the input space $D$, which in the finite-dimensional linear case coincides with the feature space.

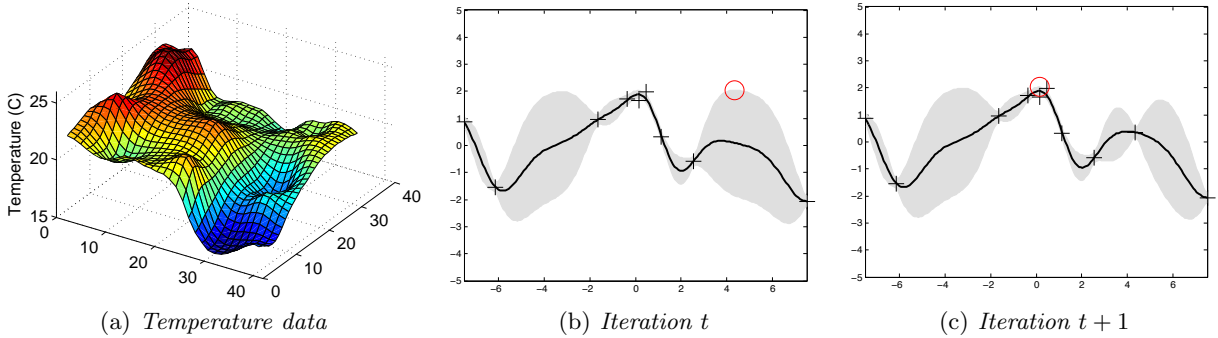

*Figure 2.* (a) Example of temperature data collected by a network of 46 sensors at Intel Research Berkeley. (b,c) Two iterations of the GP-UCB algorithm. It samples points that are either uncertain (b) or have high posterior mean (c).

Importantly, note that the information gain is a problem dependent quantity — properties of both the kernel and the input space will determine the growth of regret. In Section 5, we provide general methods for bounding $\gamma_T$, either by efficient auxiliary computations or by direct expressions for specific kernels of interest. Our results match known lower bounds (up to log factors) in both the $K$-armed bandit and the $d$-dimensional linear optimization case.

**Bounds for a GP Prior.** For finite $D$, we obtain the following bound.

**Theorem 1** *Let $\delta \in (0,1)$ and $\beta_t = 2\log(|D|t^2\pi^2/6\delta)$. Running GP-UCB with $\beta_t$ for a sample $f$ of a GP with mean function zero and covariance function $k(\boldsymbol{x}, \boldsymbol{x}')$, we obtain a regret bound of $\mathcal{O}^*(\sqrt{T\gamma_T \log|D|})$ with high probability. Precisely,*

$$\Pr\left\{R_T \leq \sqrt{C_1 T \beta_T \gamma_T} \quad \forall T \geq 1\right\} \geq 1 - \delta.$$

*where $C_1 = 8/\log(1+\sigma^{-2})$.*

The proof methodology follows Dani et al. (2007) in that we relate the regret to the growth of the log volume of the confidence ellipsoid — a novelty in our proof is showing how this growth is characterized by the information gain.

This theorem shows that, with high probability over samples from the GP, the cumulative regret is bounded in terms of the maximum information gain, forging a novel connection between GP optimization and experimental design. This link is of fundamental technical importance, allowing us to generalize Theorem 1 to infinite decision spaces. Moreover, the submodularity of $I(\boldsymbol{y}_A; \boldsymbol{f}_A)$ allows us to derive sharp a priori bounds, depending on choice and parameterization of $k$ (see Section 5). In the following theorem, we generalize our result to any compact and convex $D \subset \mathbb{R}^d$ under mild assumptions on the kernel function $k$.

**Theorem 2** *Let $D \subset [0,r]^d$ be compact and convex, $d \in \mathbb{N}$, $r > 0$. Suppose that the kernel $k(\boldsymbol{x}, \boldsymbol{x}')$ satisfies the following high probability bound on the derivatives of GP sample paths $f$: for some constants $a, b > 0$,*

$$\Pr\left\{\sup_{\boldsymbol{x} \in D} |\partial f/\partial x_j| > L\right\} \leq a e^{-(L/b)^2}, \quad j = 1, \ldots, d.$$

*Pick $\delta \in (0,1)$, and define*

$$\beta_t = 2\log(t^2 2\pi^2/(3\delta)) + 2d\log\left(t^2 dbr\sqrt{\log(4da/\delta)}\right).$$

*Running the GP-UCB with $\beta_t$ for a sample $f$ of a GP with mean function zero and covariance function $k(\boldsymbol{x}, \boldsymbol{x}')$, we obtain a regret bound of $\mathcal{O}^*(\sqrt{dT\gamma_T})$ with high probability. Precisely, with $C_1 = 8/\log(1+\sigma^{-2})$ we have*

$$\Pr\left\{R_T \leq \sqrt{C_1 T \beta_T \gamma_T} + 2 \quad \forall T \geq 1\right\} \geq 1 - \delta.$$

The main challenge in our proof (provided in the Appendix) is to lift the regret bound in terms of the confidence ellipsoid to general $D$. The smoothness assumption on $k(\boldsymbol{x}, \boldsymbol{x}')$ disqualifies GPs with highly erratic sample paths. It holds for stationary kernels $k(\boldsymbol{x}, \boldsymbol{x}') = k(\boldsymbol{x} - \boldsymbol{x}')$ which are four times differentiable (Theorem 5 of Ghosal & Roy (2006)), such as the Squared Exponential and Matérn kernels with $\nu > 2$ (see Section 5.2), while it is violated for the Ornstein-Uhlenbeck kernel (Matérn with $\nu = 1/2$; a stationary variant of the Wiener process). For the latter, sample paths $f$ are nondifferentiable almost everywhere with probability one and come with independent increments. We conjecture that a result of the form of Theorem 2 does not hold in this case.

**Bounds for Arbitrary $f$ in the RKHS.** Thus far, we have assumed that the target function $f$ is sampled from a GP prior and that the noise is $N(0, \sigma^2)$ with known variance $\sigma^2$. We now analyze GP-UCB in an agnostic setting, where $f$ is an arbitrary function from the RKHS corresponding to kernel $k(\boldsymbol{x}, \boldsymbol{x}')$. Moreover, we allow the noise variables $\varepsilon_t$ to be an arbitrary martingale difference sequence (meaning that $\mathbb{E}[\varepsilon_t \mid \boldsymbol{\varepsilon}_{<t}] = 0$ for all $t \in \mathbb{N}$), uniformly bounded by $\sigma$. Note that we still run the same GP-UCB algorithm, whose prior and noise model are misspecified in this case. Our following result shows that GP-UCB attains sublinear regret even in the agnostic setting.

**Theorem 3** *Let $\delta \in (0, 1)$. Assume that the true underlying $f$ lies in the RKHS $\mathcal{H}_k(D)$ corresponding to the kernel $k(\boldsymbol{x}, \boldsymbol{x}')$, and that the noise $\varepsilon_t$ has zero mean conditioned on the history and is bounded by $\sigma$ almost surely. In particular, assume $\|f\|_k^2 \leq B$ and let $\beta_t = 2B + 300 \gamma_t \log^3(t/\delta)$. Running GP-UCB with $\beta_t$, prior $GP(0, k(\boldsymbol{x}, \boldsymbol{x}'))$ and noise model $N(0, \sigma^2)$, we obtain a regret bound of $\mathcal{O}^*(\sqrt{T}(B\sqrt{\gamma_T} + \gamma_T))$ with high probability (over the noise). Precisely,*

$$\Pr\left\{R_T \leq \sqrt{C_1 T \beta_T \gamma_T} \quad \forall T \geq 1\right\} \geq 1 - \delta,$$

*where $C_1 = 8/\log(1 + \sigma^{-2})$.*

Note that while our theorem implicitly assumes that GP-UCB has knowledge of an upper bound on $\|f\|_k$, standard guess-and-doubling approaches suffice if no such bound is known a priori. Comparing Theorem 2 and Theorem 3, the latter holds uniformly over all functions $f$ with $\|f\|_k < \infty$, while the former is a probabilistic statement requiring knowledge of the GP that $f$ is sampled from. In contrast, if $f \sim GP(0, k(\boldsymbol{x}, \boldsymbol{x}'))$, then $\|f\|_k = \infty$ almost surely (Wahba, 1990): sample paths are rougher than RKHS functions. Neither Theorem 2 nor 3 encompasses the other.

## 5. Bounding the Information Gain

Since the bounds developed in Section 4 depend on the information gain, the key remaining question is how to bound the quantity $\gamma_T$ for practical classes of kernels.

### 5.1. Submodularity and Greedy Maximization

In order to bound $\gamma_T$, we have to maximize the information gain $F(A) = I(\boldsymbol{y}_A; f)$ over all subsets $A \subset D$ of size $T$: a combinatorial problem in general. However, as noted in Section 2, $F(A)$ is a submodular function, which implies the performance guarantee (5) for maximizing $F$ sequentially by the greedy ED rule (4). Dividing both sides of (5) by $1 - 1/e$, we can upper-bound $\gamma_T$ by $(1 - 1/e)^{-1} I(\boldsymbol{y}_{A_T}; f)$, where $A_T$ is constructed by the greedy procedure. Thus, somewhat counterintuitively, instead of using submodularity to prove that $F(A_T)$ is near-optimal, we use it in order to show that

$\gamma_T$ is "near-greedy". As noted in Section 2, the ED rule does not depend on observations $y_t$ and can be run without evaluating $f$.

The importance of this greedy bound is twofold. First, it allows us to numerically compute highly problem-specific bounds on $\gamma_T$, which can be plugged into our results in Section 4 to obtain high-probability bounds on $R_T$. This being a laborious procedure, one would prefer *a priori* bounds for $\gamma_T$ in practice which are simple analytical expressions of $T$ and parameters of $k$. In this section, we sketch a general procedure for obtaining such expressions, instantiating them for a number of commonly used covariance functions, once more relying crucially on the greedy ED rule upper bound. Suppose that $D$ is finite for now, and let $\boldsymbol{f} = [f(\boldsymbol{x})]_{\boldsymbol{x} \in D}$, $\boldsymbol{K}_D = [k(\boldsymbol{x}, \boldsymbol{x}')]_{\boldsymbol{x}, \boldsymbol{x}' \in D}$. Sampling $f$ at $\boldsymbol{x}_t$, we obtain $y_t \sim N(\boldsymbol{v}_t^T \boldsymbol{f}, \sigma^2)$, where $\boldsymbol{v}_t \in \mathbb{R}^{|D|}$ is the indicator vector associated with $\boldsymbol{x}_t$. We can upper-bound the greedy maximum once more, by relaxing this constraint to $\|\boldsymbol{v}_t\| = 1$ in round $t$ of the sequential method. For this relaxed greedy procedure, all $\boldsymbol{v}_t$ are leading eigenvectors of $\boldsymbol{K}_D$, since successive covariance matrices of $P(\boldsymbol{f}|\boldsymbol{y}_{t-1})$ share their eigenbasis with $\boldsymbol{K}_D$, while eigenvalues are damped according to how many times the corresponding eigenvector is selected. We can upper-bound the information gain by considering the worst-case allocation of $T$ samples to the $\min\{T, |D|\}$ leading eigenvectors of $\boldsymbol{K}_D$:

$$\gamma_T \leq \frac{1/2}{1 - e^{-1}} \max_{(m_t)} \sum_{t=1}^{|D|} \log(1 + \sigma^{-2} m_t \hat{\lambda}_t), \quad (8)$$

subject to $\sum_t m_t = T$, and $\text{spec}(\boldsymbol{K}_D) = \{\hat{\lambda}_1 \geq \hat{\lambda}_2 \geq \ldots\}$. We can split the sum into two parts in order to obtain a bound to leading order. The following Theorem captures this intuition:

**Theorem 4** *For any $T \in \mathbb{N}$ and any $T_* = 1, \ldots, T$:*

$$\gamma_T \leq \mathcal{O}\big(\sigma^{-2}[B(T_*)T + T_*(\log n_T T)]\big),$$

*where $n_T = \sum_{t=1}^{|D|} \hat{\lambda}_t$ and $B(T_*) = \sum_{t=T_*+1}^{|D|} \hat{\lambda}_t$.*

Therefore, if for some $T_* = o(T)$ the first $T_*$ eigenvalues carry most of the total mass $n_T$, the information gain will be small. The more rapidly the spectrum of $\boldsymbol{K}_D$ decays, the slower the growth of $\gamma_T$. Figure 3 illustrates this intuition.

### 5.2. Bounds for Common Kernels

In this section we bound $\gamma_T$ for a range of commonly used covariance functions: finite dimensional linear, Squared Exponential and Matérn kernels. Together with our results in Section 4, these imply sublinear regret bounds for GP-UCB in all cases.

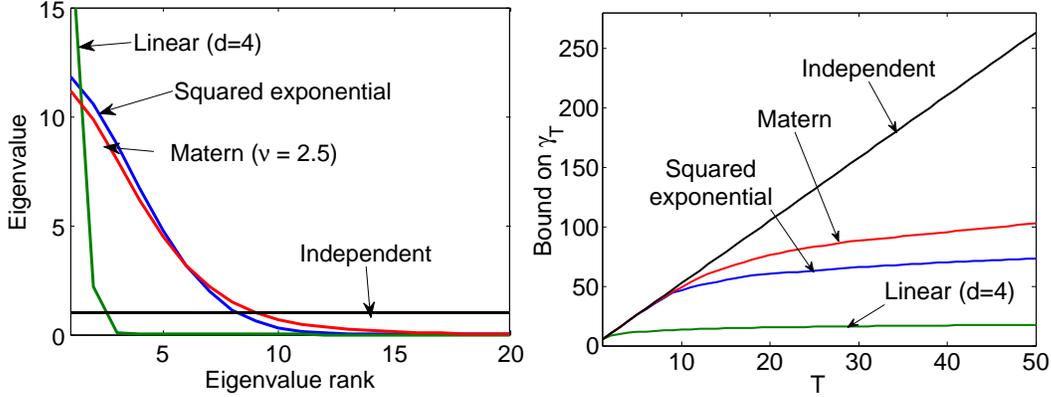

*Figure 3.* Spectral decay (left) and information gain bound (right) for independent (diagonal), linear, squared exponential and Matérn kernels ($\nu = 2.5$.) with equal trace.

*Finite dimensional linear* kernels have the form $k(\boldsymbol{x}, \boldsymbol{x}') = \boldsymbol{x}^T \boldsymbol{x}'$. GPs with this kernel correspond to random linear functions $f(\boldsymbol{x}) = \boldsymbol{w}^T \boldsymbol{x}$, $\boldsymbol{w} \sim N(\boldsymbol{0}, \boldsymbol{I})$.

The *Squared Exponential kernel* is $k(\boldsymbol{x}, \boldsymbol{x}') = \exp(-(2l^2)^{-1}\|\boldsymbol{x} - \boldsymbol{x}'\|^2)$, $l$ a lengthscale parameter. Sample functions are differentiable to any order almost surely (Rasmussen & Williams, 2006).

The *Matérn kernel* is given by $k(\boldsymbol{x}, \boldsymbol{x}') = (2^{1-\nu}/\Gamma(\nu))r^\nu B_\nu(r)$, $r = (\sqrt{2\nu}/l)\|\boldsymbol{x} - \boldsymbol{x}'\|$, where $\nu$ controls the smoothness of sample paths (the smaller, the rougher) and $B_\nu$ is a modified Bessel function. Note that as $\nu \to \infty$, appropriately rescaled Matérn kernels converge to the Squared Exponential kernel.

Figure 4 shows random functions drawn from GP distributions with the above kernels.

**Theorem 5** *Let $D \subset \mathbb{R}^d$ be compact and convex, $d \in \mathbb{N}$. Assume the kernel function satisfies $k(\boldsymbol{x}, \boldsymbol{x}') \leq 1$.*

1. *Finite spectrum. For the d-dimensional Bayesian linear regression case: $\gamma_T = \mathcal{O}(d \log T)$.*
2. *Exponential spectral decay. For the Squared Exponential kernel: $\gamma_T = \mathcal{O}((\log T)^{d+1})$.*
3. *Power law spectral decay. For Matérn kernels with $\nu > 1$: $\gamma_T = \mathcal{O}(T^{d(d+1)/(2\nu+d(d+1))}(\log T))$.*

A proof of Theorem 5 is given in the Appendix, , we only sketch the idea here. $\gamma_T$ is bounded by Theorem 4 in terms the eigendecay of the kernel matrix $\boldsymbol{K}_D$. If $D$ is infinite or very large, we can use the operator spectrum of $k(\boldsymbol{x}, \boldsymbol{x}')$, which likewise decays rapidly. For the kernels of interest here, asymptotic expressions for the operator eigenvalues are given in Seeger et al. (2008), who derived bounds on the information gain for fixed and random designs (in contrast to the worst-case information gain considered here, which is substantially more challenging to bound). The main challenge in the proof is to ensure the existence of discretizations $D_T \subset D$, dense in the limit, for which tail sums $B(T_*)/n_T$ in Theorem 4 are close to corresponding operator spectra tail sums.

Together with Theorems 2 and 3, this result guarantees sublinear regret of GP-UCB for any dimension (see Figure 1). For the Squared Exponential kernel, the dimension $d$ appears as exponent of $\log T$ only, so that the regret grows at most as $\mathcal{O}^*(\sqrt{T}(\log T)^{\frac{d+1}{2}})$ – the high degree of smoothness of the sample paths effectively combats the curse of dimensionality.

## 6. Experiments

We compare GP-UCB with heuristics such as the Expected Improvement (EI) and Most Probable Improvement (MPI), and with naive methods which choose points of maximum mean or variance only, both on synthetic and real sensor network data.

For synthetic data, we sample random functions from a squared exponential kernel with lengthscale parameter 0.2. The sampling noise variance $\sigma^2$ was set to 0.025 or 5% of the signal variance. Our decision set $D = [0, 1]$ is uniformly discretized into 1000 points. We run each algorithm for $T = 1000$ iterations with $\delta = 0.1$, averaging over 30 trials (samples from the kernel). While the choice of $\beta_t$ as recommended by Theorem 1 leads to competitive performance of GP-UCB, we find (using cross-validation) that the algorithm is improved by scaling $\beta_t$ down by a factor 5. Note that we did not optimize constants in our regret bounds.

Next, we use temperature data collected from 46 sensors deployed at Intel Research Berkeley over 5 days at 1 minute intervals, pertaining to the example in Section 2. We take the first two-thirds of the data set to compute the empirical covariance of the sensor readings, and use it as the kernel matrix. The functions $f$ for optimization consist of one set of observations from all the sensors taken from the remaining third of the data set, and the results (for $T = 46$, $\sigma^2 = 0.5$ or 5% noise, $\delta = 0.1$) were averaged over 2000 possible choices of the objective function.

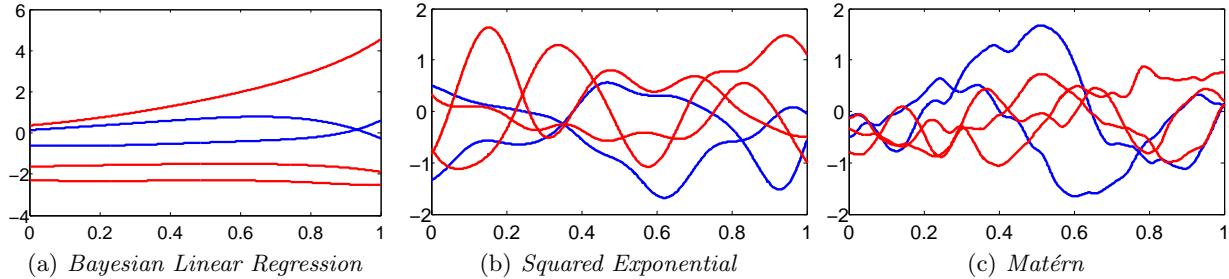

(a) *Bayesian Linear Regression*    (b) *Squared Exponential*    (c) *Matérn*

Figure 4. Sample functions drawn from a GP with linear, squared exponential and Matérn kernels ($\nu = 2.5$.)

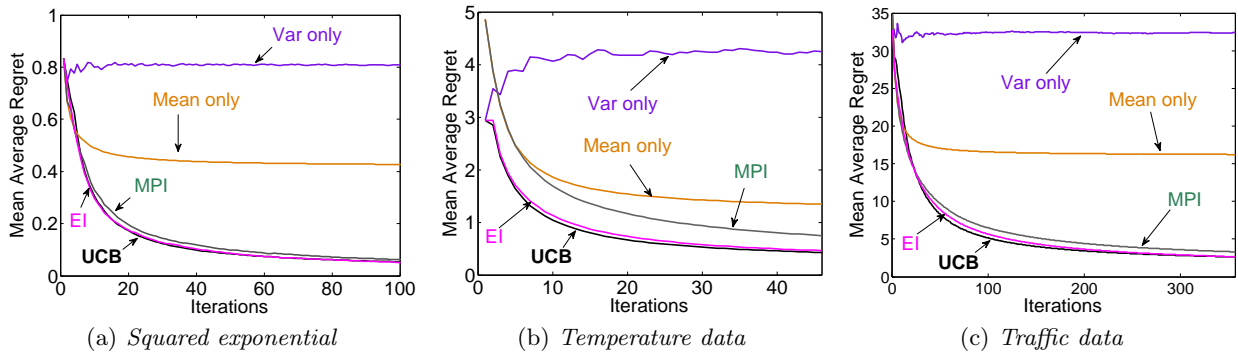

(a) *Squared exponential*    (b) *Temperature data*    (c) *Traffic data*

Figure 5. Comparison of performance: GP-UCB and various heuristics on synthetic (a), and sensor network data (b, c).

Lastly, we take data from traffic sensors deployed along the highway I-880 South in California. The goal was to find the point of minimum speed in order to identify the most congested portion of the highway; we used traffic speed data for all working days from 6 AM to 11 AM for one month, from 357 sensors. We again use the covariance matrix from two-thirds of the data set as kernel matrix, and test on the other third. The results (for $T = 357$, $\sigma^2 = 4.78$ or 5% noise, $\delta = 0.1$) were averaged over 900 runs.

Figure 5 compares the mean average regret incurred by the different heuristics and the GP-UCB algorithm on synthetic and real data. For temperature data, the GP-UCB algorithm and EI heuristic clearly outperform the others, and do not exhibit significant difference between each other. On synthetic and traffic data MPI does equally well. In summary, GP-UCB performs at least on par with the existing approaches which are not equipped with regret bounds.

## 7. Conclusions

We prove the first sublinear regret bounds for GP optimization with commonly used kernels (see Figure 1), both for $f$ sampled from a known GP and $f$ of low RKHS norm. We analyze GP-UCB, an intuitive, Bayesian upper confidence bound based sampling rule. Our regret bounds crucially depend on the information gain due to sampling, establishing a novel connection between bandit optimization and experimental design. We bound the information gain in terms of the kernel spectrum, providing a general methodology for obtaining regret bounds with kernels of interest. Our experiments on real sensor network data indicate that GP-UCB performs at least on par with competing criteria for GP optimization, for which no regret bounds are known at present. Our results provide an interesting step towards understanding exploration–exploitation tradeoffs with complex utility functions.

## Acknowledgements

We thank Marcus Hutter for insightful comments on an earlier version of this paper. This research was partially supported by ONR grant N00014-09-1-1044, NSF grant CNS-0932392, a gift from Microsoft Corporation and the Excellence Initiative of the German research foundation (DFG).

# A. Regret Bounds for Target Function Sampled from GP

In this section, we provide details for the proofs of Theorem 1 and Theorem 2. In both cases, the strategy is to show that $|f(\boldsymbol{x}) - \mu_{t-1}(\boldsymbol{x})| \leq \beta_t^{1/2} \sigma_{t-1}(\boldsymbol{x})$ for all $t \in \mathbb{N}$ and all $\boldsymbol{x} \in D$, or in the infinite case, all $\boldsymbol{x}$ in a discretization of $D$ which becomes dense as $t$ gets large.

## A.1. Finite Decision Set

We begin with the finite case, $|D| < \infty$.

**Lemma 5.1** *Pick* $\delta \in (0,1)$ *and set* $\beta_t = 2\log(|D|\pi_t/\delta)$, *where* $\sum_{t \geq 1} \pi_t^{-1} = 1$, $\pi_t > 0$. *Then,*

$$|f(\boldsymbol{x}) - \mu_{t-1}(\boldsymbol{x})| \leq \beta_t^{1/2} \sigma_{t-1}(\boldsymbol{x}) \quad \forall \boldsymbol{x} \in D \, \forall t \geq 1$$

*holds with probability* $\geq 1 - \delta$.

**Proof** Fix $t \geq 1$ and $\boldsymbol{x} \in D$. Conditioned on $\boldsymbol{y}_{t-1} = (y_1, \ldots, y_{t-1})$, $\{\boldsymbol{x}_1, \ldots, \boldsymbol{x}_{t-1}\}$ are deterministic, and $f(\boldsymbol{x}) \sim N(\mu_{t-1}(\boldsymbol{x}), \sigma_{t-1}^2(\boldsymbol{x}))$. Now, if $r \sim N(0,1)$, then

$$\Pr\{r > c\} = e^{-c^2/2}(2\pi)^{-1/2} \int e^{-(r-c)^2/2 - c(r-c)} \, dr$$
$$\leq e^{-c^2/2} \Pr\{r > 0\} = (1/2)e^{-c^2/2}$$

for $c > 0$, since $e^{-c(r-c)} \leq 1$ for $r \geq c$. Therefore, $\Pr\{|f(\boldsymbol{x}) - \mu_{t-1}(\boldsymbol{x})| > \beta_t^{1/2}\sigma_{t-1}(\boldsymbol{x})\} \leq e^{-\beta_t/2}$, using $r = (f(\boldsymbol{x}) - \mu_{t-1}(\boldsymbol{x}))/\sigma_{t-1}(\boldsymbol{x})$ and $c = \beta_t^{1/2}$. Applying the union bound,

$$|f(\boldsymbol{x}) - \mu_{t-1}(\boldsymbol{x})| \leq \beta_t^{1/2}\sigma_{t-1}(\boldsymbol{x}) \quad \forall \boldsymbol{x} \in D$$

holds with probability $\geq 1 - |D|e^{-\beta_t/2}$. Choosing $|D|e^{-\beta_t/2} = \delta/\pi_t$ and using the union bound for $t \in \mathbb{N}$, the statement holds. For example, we can use $\pi_t = \pi^2 t^2/6$. ∎

**Lemma 5.2** *Fix $t \geq 1$. If $|f(\boldsymbol{x}) - \mu_{t-1}(\boldsymbol{x})| \leq \beta_t^{1/2}\sigma_{t-1}(\boldsymbol{x})$ for all $\boldsymbol{x} \in D$, then the regret $r_t$ is bounded by $2\beta_t^{1/2}\sigma_{t-1}(\boldsymbol{x}_t)$.*

**Proof** By definition of $\boldsymbol{x}_t$: $\mu_{t-1}(\boldsymbol{x}_t) + \beta_t^{1/2}\sigma_{t-1}(\boldsymbol{x}_t) \geq \mu_{t-1}(\boldsymbol{x}^*) + \beta_t^{1/2}\sigma_{t-1}(\boldsymbol{x}^*) \geq f(\boldsymbol{x}^*)$. Therefore,

$$r_t = f(\boldsymbol{x}^*) - f(\boldsymbol{x}_t) \leq \beta_t^{1/2}\sigma_{t-1}(\boldsymbol{x}_t) + \mu_{t-1}(\boldsymbol{x}_t) - f(\boldsymbol{x}_t)$$
$$\leq 2\beta_t^{1/2}\sigma_{t-1}(\boldsymbol{x}_t).$$
∎

**Lemma 5.3** *The information gain for the points selected can be expressed in terms of the predictive variances. If $\boldsymbol{f}_T = (f(\boldsymbol{x}_t)) \in \mathbb{R}^T$:*

$$\mathrm{I}(\boldsymbol{y}_T; \boldsymbol{f}_T) = \frac{1}{2} \sum_{t=1}^T \log\left(1 + \sigma^{-2}\sigma_{t-1}^2(\boldsymbol{x}_t)\right).$$

**Proof** Recall that $\mathrm{I}(\boldsymbol{y}_T; \boldsymbol{f}_T) = \mathrm{H}(\boldsymbol{y}_T) - (1/2)\log|2\pi e \sigma^2 \boldsymbol{I}|$. Now, $\mathrm{H}(\boldsymbol{y}_T) = \mathrm{H}(\boldsymbol{y}_{T-1}) + \mathrm{H}(y_T|\boldsymbol{y}_{T-1}) = \mathrm{H}(\boldsymbol{y}_{T-1}) + \log(2\pi e(\sigma^2 + \sigma_{T-1}^2(\boldsymbol{x}_T)))/2$. Here, we use that $\boldsymbol{x}_1, \ldots, \boldsymbol{x}_T$ are deterministic conditioned on $\boldsymbol{y}_{T-1}$, and that the conditional variance $\sigma_{T-1}^2(\boldsymbol{x}_T)$ does not depend on $\boldsymbol{y}_{T-1}$. The result follows by induction. ∎

**Lemma 5.4** *Pick $\delta \in (0, 1)$ and let $\beta_t$ be defined as in Lemma 5.1. Then, the following holds with probability $\geq 1 - \delta$:*

$$\sum_{t=1}^T r_t^2 \leq \beta_T C_1 \mathrm{I}(\boldsymbol{y}_T; \boldsymbol{f}_T) \leq C_1 \beta_T \gamma_T \quad \forall T \geq 1,$$

where $C_1 := 8/\log(1 + \sigma^{-2}) \geq 8\sigma^2$.

**Proof** By Lemma 5.1 and Lemma 5.2, we have that $\{r_t^2 \leq 4\beta_t \sigma_{t-1}^2(\boldsymbol{x}_t) \,\forall t \geq 1\}$ with probability $\geq 1 - \delta$. Now, $\beta_t$ is nondecreasing, so that

$$4\beta_t \sigma_{t-1}^2(\boldsymbol{x}_t) \leq 4\beta_T \sigma^2 (\sigma^{-2}\sigma_{t-1}^2(\boldsymbol{x}_t))$$
$$\leq 4\beta_T \sigma^2 C_2 \log(1 + \sigma^{-2}\sigma_{t-1}^2(\boldsymbol{x}_t))$$

with $C_2 = \sigma^{-2}/\log(1 + \sigma^{-2}) \geq 1$, since $s^2 \leq C_2 \log(1 + s^2)$ for $s \in [0, \sigma^{-2}]$, and $\sigma^{-2}\sigma_{t-1}^2(\boldsymbol{x}_t) \leq \sigma^{-2}k(\boldsymbol{x}_t, \boldsymbol{x}_t) \leq \sigma^{-2}$. Noting that $C_1 = 8\sigma^2 C_2$, the result follows by plugging in the representation of Lemma 5.3. ∎

Finally, Theorem 1 is a simple consequence of Lemma 5.4, since $R_T^2 \leq T \sum_{t=1}^T r_t^2$ by the Cauchy-Schwarz inequality.

### A.2. General Decision Set

Theorem 2 extends the statement of Theorem 1 to the general case of $D \subset \mathbb{R}^d$ compact. We cannot expect this generalization to work without any assumptions on the kernel $k(\boldsymbol{x}, \boldsymbol{x}')$. For example, if $k(\boldsymbol{x}, \boldsymbol{x}') = e^{-\|\boldsymbol{x}-\boldsymbol{x}'\|}$ (Ornstein-Uhlenbeck), while sample paths $f$ are a.s. continuous, they are still very erratic: $f$ is a.s. nondifferentiable almost everywhere, and the process comes with independent increments, a stationary variant of Brownian motion. The additional assumption on $k$ in Theorem 2 is rather mild and is satisfied by several common kernels, as discussed in Section 4.

Recall that the finite case proof is based on Lemma 5.1 paving the way for Lemma 5.2. However, Lemma 5.1 does not hold for infinite $D$. First, let us observe that we have confidence on all decisions actually chosen.

**Lemma 5.5** *Pick $\delta \in (0, 1)$ and set $\beta_t = 2\log(\pi_t/\delta)$, where $\sum_{t \geq 1} \pi_t^{-1} = 1$, $\pi_t > 0$. Then,*

$$|f(\boldsymbol{x}_t) - \mu_{t-1}(\boldsymbol{x}_t)| \leq \beta_t^{1/2}\sigma_{t-1}(\boldsymbol{x}_t) \quad \forall t \geq 1$$

*holds with probability $\geq 1 - \delta$.*

**Proof** Fix $t \geq 1$ and $\boldsymbol{x} \in D$. Conditioned on $\boldsymbol{y}_{t-1} = (y_1, \ldots, y_{t-1})$, $\{\boldsymbol{x}_1, \ldots, \boldsymbol{x}_{t-1}\}$ are deterministic, and $f(\boldsymbol{x}) \sim N(\mu_{t-1}(\boldsymbol{x}), \sigma_{t-1}^2(\boldsymbol{x}))$. As before, $\Pr\{|f(\boldsymbol{x}_t) - \mu_{t-1}(\boldsymbol{x}_t)| > \beta_t^{1/2}\sigma_{t-1}(\boldsymbol{x}_t)\} \leq e^{-\beta_t/2}$. Since $e^{-\beta_t/2} = \delta/\pi_t$ and using the union bound for $t \in \mathbb{N}$, the statement holds. ∎

Purely for the sake of analysis, we use a set of discretizations $D_t \subset D$, where $D_t$ will be used at time

$t$ in the analysis. Essentially, we use this to obtain a valid confidence interval on $\boldsymbol{x}^*$. The following lemma provides a confidence bound for these subsets.

**Lemma 5.6** *Pick $\delta \in (0,1)$ and set $\beta_t = 2\log(|D_t|\pi_t/\delta)$, where $\sum_{t\geq 1}\pi_t^{-1} = 1$, $\pi_t > 0$. Then,*

$$|f(\boldsymbol{x}) - \mu_{t-1}(\boldsymbol{x})| \leq \beta_t^{1/2}\sigma_{t-1}(\boldsymbol{x}) \quad \forall \boldsymbol{x} \in D_t, \forall t \geq 1$$

*holds with probability $\geq 1 - \delta$.*

**Proof** The proof is identical to that in Lemma 5.1, except now we use $D_t$ at each timestep. ∎

Now by assumption and the union bound, we have that

$$\Pr\left\{\forall j, \forall \boldsymbol{x} \in D, |\partial f/(\partial x_j)| < L\right\} \geq 1 - dae^{-L^2/b^2}.$$

which implies that, with probability greater than $1 - dae^{-L^2/b^2}$, we have that

$$\forall \boldsymbol{x} \in D, |f(x) - f(x')| \leq L\|x - x'\|_1. \qquad (9)$$

This allows us to obtain confidence on $\boldsymbol{x}^\star$ as follows.

Now let us choose a discretization $D_t$ of size $(\tau_t)^d$ so that for all $\boldsymbol{x} \in D_t$

$$\|\boldsymbol{x} - [\boldsymbol{x}]_t\|_1 \leq rd/\tau_t$$

where $[\boldsymbol{x}]_t$ denotes the closest point in $D_t$ to $\boldsymbol{x}$. A sufficient discretization has each coordinate with $\tau_t$ uniformly spaced points.

**Lemma 5.7** *Pick $\delta \in (0,1)$ and set $\beta_t = 2\log(2\pi_t/\delta) + 4d\log(dtbr\sqrt{\log(2da/\delta)})$, where $\sum_{t\geq 1}\pi_t^{-1} = 1$, $\pi_t > 0$. Let $\tau_t = dt^2br\sqrt{\log(2da/\delta)}$ Let $[\boldsymbol{x}^*]_t$ denotes the closest point in $D_t$ to $\boldsymbol{x}^*$. Hence, Then,*

$$|f(\boldsymbol{x}^*) - \mu_{t-1}([\boldsymbol{x}^*]_t)| \leq \beta_t^{1/2}\sigma_{t-1}([\boldsymbol{x}^*]_t) + \frac{1}{t^2} \quad \forall t \geq 1$$

*holds with probability $\geq 1 - \delta$.*

**Proof** Using (9), we have that with probability greater than $1 - \delta/2$,

$$\forall \boldsymbol{x} \in D, |f(x) - f(x')| \leq b\sqrt{\log(2da/\delta)}\|x - x'\|_1.$$

Hence,

$$\forall \boldsymbol{x} \in D_t, |f(x) - f([x]_t)| \leq rdb\sqrt{\log(2da/\delta)}/\tau_t.$$

Now by choosing $\tau_t = dt^2br\sqrt{\log(2da/\delta)}$, we have that

$$\forall \boldsymbol{x} \in D_t, |f(x) - f([x]_t)| \leq \frac{1}{t^2}$$

This implies that $|D_t| = (dt^2br\sqrt{\log(2da/\delta)})^d$. Using $\delta/2$ in Lemma 5.6, we can apply the confidence bound to $[\boldsymbol{x}^*]_t$ (as this lives in $D_t$) to obtain the result. ∎

Now we are able to bound the regret.

**Lemma 5.8** *Pick $\delta \in (0,1)$ and set $\beta_t = 2\log(4\pi_t/\delta) + 4d\log(dtbr\sqrt{\log(4da/\delta)})$, where $\sum_{t\geq 1}\pi_t^{-1} = 1$, $\pi_t > 0$. Then, with probability greater than $1 - \delta$, for all $t \in \mathbb{N}$, the regret is bounded as follows:*

$$r_t \leq 2\beta_t^{1/2}\sigma_{t-1}(\boldsymbol{x}_t) + \frac{1}{t^2}.$$

**Proof** We use $\delta/2$ in both Lemma 5.5 and Lemma 5.7, so that these events hold with probability greater than $1 - \delta$. Note that the specification of $\beta_t$ in the above lemma is greater than the specification used in Lemma 5.5 (with $\delta/2$), so this choice is valid.

By definition of $\boldsymbol{x}_t$: $\mu_{t-1}(\boldsymbol{x}_t) + \beta_t^{1/2}\sigma_{t-1}(\boldsymbol{x}_t) \geq \mu_{t-1}([\boldsymbol{x}^*]_t) + \beta_t^{1/2}\sigma_{t-1}([\boldsymbol{x}^*]_t)$. Also, by Lemma 5.7, we have that $\mu_{t-1}([\boldsymbol{x}^*]_t) + \beta_t^{1/2}\sigma_{t-1}([\boldsymbol{x}^*]_t) + 1/t^2 \geq f(\boldsymbol{x}^*)$, which implies $\mu_{t-1}(\boldsymbol{x}_t) + \beta_t^{1/2}\sigma_{t-1}(\boldsymbol{x}_t) \geq f(\boldsymbol{x}^*) - 1/t^2$. Therefore,

$$\begin{aligned}r_t &= f(\boldsymbol{x}^*) - f(\boldsymbol{x}_t) \\ &\leq \beta_t^{1/2}\sigma_{t-1}(\boldsymbol{x}_t) + 1/t^2 + \mu_{t-1}(\boldsymbol{x}_t) - f(\boldsymbol{x}_t) \\ &\leq 2\beta_t^{1/2}\sigma_{t-1}(\boldsymbol{x}_t) + 1/t^2.\end{aligned}$$

which completes the proof. ∎

Now we are ready to complete the proof of Theorem 2. As shown in the proof of Lemma 5.4, we have that with probability greater than $1 - \delta$,

$$\sum_{t=1}^T 4\beta_t\sigma_{t-1}^2(\boldsymbol{x}_t) \leq C_1\beta_T\gamma_T \quad \forall T \geq 1,$$

so that by Cauchy-Schwarz:

$$\sum_{t=1}^T 2\beta_t^{1/2}\sigma_{t-1}(\boldsymbol{x}_t) \leq \sqrt{C_1 T \beta_T \gamma_T} \quad \forall T \geq 1,$$

Hence,

$$\sum_{t=1}^T r_t \leq \sqrt{C_1 T \beta_T \gamma_T} + \pi^2/6 \quad \forall T \geq 1,$$

(since $\sum 1/t^2 = \pi^2/6$). Theorem 2 now follows.

Finally, we now discuss the additional assumption on $k$ in Theorem 2. For samples $f$ of the GP, consider partial derivatives $\partial f/(\partial x_j)$ of this sample path for $j = 1, \ldots, d$. Theorem 5 of Ghosal & Roy (2006) states that if derivatives up to fourth order exists for $(\boldsymbol{x}, \boldsymbol{x}') \mapsto k(\boldsymbol{x}, \boldsymbol{x}')$, then $f$ is almost surely continuously differentiable, with $\partial f/(\partial x_j)$ distributed as Gaussian processes again. Moreover, there are constants $a, b_j > 0$ such that

$$\Pr\left\{\sup_{\boldsymbol{x} \in D} |\partial f/(\partial x_j)| > L\right\} \leq a e^{-b_j L^2}. \qquad (10)$$

Picking $L = [\log(da2/\delta)/\min_j b_j]^{1/2}$, we have that $ae^{-b_j L^2} \leq \delta/(2d)$ for all $j = 1, \ldots, d$, so that for $K_1 = d^{1/2}L$, by the mean value theorem, we have $\Pr\{|f(\boldsymbol{x}) - f(\boldsymbol{x}')| \leq K_1\|\boldsymbol{x} - \boldsymbol{x}'\| \,\forall\, \boldsymbol{x}, \boldsymbol{x}' \in D\} \geq 1 - \delta/2$. Also, note that $K_1 = \mathcal{O}((\log \delta^{-1})^{1/2})$.

This statement is about the joint distribution of $f(\cdot)$ and its partial derivatives w.r.t. each component. For a certain event in this sample space, all $\partial f/(\partial x_j)$ exist, are continuous, and the complement of (10) holds for all $j$. Theorem 5 of Ghosal & Roy (2006), together with the union bound, implies that this event has probability $\geq 1 - \delta/2$. Derivatives up to fourth order exist for the Gaussian covariance function, and for Matérn kernels with $\nu > 2$ (Stein, 1999).

## B. Regret Bound for Target Function in RKHS

In this section, we detail a proof of Theorem 3. Recall that in this setting, we do not know the generator of the target function $f$, but only a bound on its RKHS norm $\|f\|_k$.

Recall the posterior mean function $\mu_T(\cdot)$ and posterior covariance function $k_T(\cdot, \cdot)$ from Section 2, conditioned on data $(\boldsymbol{x}_t, y_t)$, $t = 1, \ldots, T$. It is easy to see that the RKHS norm corresponding to $k_T$ is given by

$$\|f\|_{k_T}^2 = \|f\|_k^2 + \sigma^{-2} \sum_{t=1}^T f(\boldsymbol{x}_t)^2.$$

This implies that $\mathcal{H}_k(D) = \mathcal{H}_{k_T}(D)$ for any $T$, while the RKHS inner products are different: $\|f\|_{k_T} \geq \|f\|_k$. Since $\langle f(\cdot), k_T(\cdot, \boldsymbol{x})\rangle_{k_T} = f(\boldsymbol{x})$ for any $f \in \mathcal{H}_{k_T}(D)$ by the reproducing property, then

$$\begin{aligned}|\mu_t(\boldsymbol{x}) - f(\boldsymbol{x})| &\leq k_T(\boldsymbol{x}, \boldsymbol{x})^{1/2} \|\mu_t - f\|_{k_T} \\ &= \sigma_T(\boldsymbol{x}) \|\mu_t - f\|_{k_T}\end{aligned} \qquad (11)$$

by the Cauchy-Schwarz inequality.

Compared to our other results, Theorem 3 is an agnostic statement, in that the assumptions the Bayesian UCB algorithm bases its predictions on differ from how $f$ and data $y_t$ are generated. First, $f$ is not drawn from a GP, but can be an arbitrary function from $\mathcal{H}_k(D)$. Second, while the UCB method assumes that the noise $\varepsilon_t = y_t - f(\boldsymbol{x}_t)$ is drawn independently from $N(0, \sigma^2)$, the true sequence of noise variables $\varepsilon_t$ can be a uniformly bounded martingale difference sequence: $\varepsilon_t \leq \sigma$ for all $t \in \mathbb{N}$. All we have to do in order to lift the proof of Theorem 1 to the agnostic setting is to establish an analogue to Lemma 5.1, by way of the following concentration result.

**Theorem 6** *Let $\delta \in (0, 1)$. Assume the noise variables $\varepsilon_t$ are uniformly bounded by $\sigma$. Define:*

$$\beta_t = 2\|f\|_k^2 + 300 \gamma_t \ln^3(t/\delta),$$

*Then*

$$\Pr\left\{\forall T, \forall x \in D, \;\; |\mu_T(\boldsymbol{x}) - f(\boldsymbol{x})| \leq \beta_{T+1}^{1/2} \sigma_T(\boldsymbol{x})\right\} \geq 1 - \delta.$$

### B.1. Concentration of Martingales

In our analysis, we use the following Bernstein-type concentration inequality for martingale differences, due to Freedman (1975) (see also Theorem 3.15 of McDiarmid 1998).

**Theorem 7 (Freedman)** *Suppose $X_1, \ldots, X_T$ is a martingale difference sequence, and $b$ is an uniform upper bound on the steps $X_i$. Let $V$ denote the sum of conditional variances,*

$$V = \sum_{i=1}^n \mathbf{Var}\left(X_i \mid X_1, \ldots, X_{i-1}\right).$$

*Then, for every $a, v > 0$,*

$$\Pr\left\{\sum X_i \geq a \text{ and } V \leq v\right\} \leq \exp\left(\frac{-a^2}{2v + 2ab/3}\right).$$

### B.2. Proof of Theorem 6

We will show that:

$$\Pr\left\{\forall T, \;\; \|\mu_T - f\|_{k_T}^2 \leq \beta_{T+1}\right\} \geq 1 - \delta.$$

Theorem 6 then follows from (11). Recall that $\varepsilon_t = y_t - f(\boldsymbol{x}_t)$. We will analyze the quantity $Z_T = \|\mu_T - f\|_{k_T}^2$, measuring the error of $\mu_T$ as approximation to $f$ under the RKHS norm of $\mathcal{H}_{k_T}(D)$. The following lemma provides the connection with the information gain. This lemma is important since our concentration argument is an inductive argument — roughly speaking, we condition on getting concentration in the past, in order to achieve good concentration in the future.

**Lemma 7.1** *We have that*

$$\sum_{t=1}^T \min\{\sigma^{-2}\sigma_{t-1}^2(\boldsymbol{x}_t), \alpha\} \leq \frac{2\alpha}{\log(1+\alpha)} \gamma_T, \quad \alpha > 0.$$

**Proof** We have that $\min\{r, \alpha\} \leq (\alpha/\log(1+\alpha))\log(1+r)$. The statement follows from Lemma 5.3. ∎

The next lemma bounds the growth of $Z_T$. It is formulated in terms of normalized quantities: $\widetilde{\varepsilon}_t = \varepsilon_t/\sigma$, $\widetilde{f} = f/\sigma$, $\widetilde{\mu}_t = \mu_t/\sigma$, $\widetilde{\sigma}_t = \sigma_t/\sigma$. Also, to ease notation, we will use $\mu_{t-1}$, $\sigma_{t-1}$ as shorthand for $\mu_{t-1}(\boldsymbol{x}_t)$, $\sigma_{t-1}(\boldsymbol{x}_t)$.

**Lemma 7.2** *For all $T \in \mathbb{N}$,*

$$Z_T \leq \|f\|_k^2 + 2\sum_{t=1}^T \widetilde{\varepsilon}_t \frac{\widetilde{\mu}_{t-1} - \widetilde{f}(\boldsymbol{x}_t)}{1 + \widetilde{\sigma}_{t-1}^2}$$
$$+ \sum_{t=1}^T \widetilde{\varepsilon}_t^2 \frac{\widetilde{\sigma}_{t-1}^2}{1 + \widetilde{\sigma}_{t-1}^2}.$$

**Proof** If $\boldsymbol{\alpha}_t = (\boldsymbol{K}_t + \sigma^2 \boldsymbol{I})^{-1} \boldsymbol{y}_t$, then $\mu_t(\boldsymbol{x}) = \boldsymbol{\alpha}_t^T \boldsymbol{k}_t(\boldsymbol{x})$. Then, $\langle \mu_T, f \rangle_k = \boldsymbol{f}_T^T \boldsymbol{\alpha}_T$, $\|\mu_T\|_k^2 = \boldsymbol{y}_T^T \boldsymbol{\alpha}_T - \sigma^2 \|\boldsymbol{\alpha}_T\|^2$. Moreover, for $t \leq T$, $\mu_T(x_t) = \boldsymbol{\delta}_t^T \boldsymbol{K}_T (\boldsymbol{K}_T + \sigma^2 \boldsymbol{I})^{-1} \boldsymbol{y}_T = y_t - \sigma^2 \alpha_t$. Since $Z_T = \|\mu_T - f\|_k + \sigma^{-2} \sum_{t \leq T}(\mu_T(\boldsymbol{x}_t) - f(\boldsymbol{x}_t))^2$, we have that

$$Z_T = \|f\|_k^2 - 2\boldsymbol{f}_T^T \boldsymbol{\alpha}_T + \boldsymbol{y}_T^T \boldsymbol{\alpha}_T - \sigma^2 \|\boldsymbol{\alpha}_T\|^2$$
$$+ \sigma^{-2} \sum_{t=1}^T (\varepsilon_t - \sigma^2 \alpha_t)^2 = \|f\|_k^2$$
$$- \boldsymbol{y}_T^T (\boldsymbol{K}_T + \sigma^2 \boldsymbol{I})^{-1} \boldsymbol{y}_T + \sigma^{-2} \|\boldsymbol{\varepsilon}_T\|^2.$$

Now, $-\boldsymbol{y}_T^T(\boldsymbol{K}_T + \sigma^2 \boldsymbol{I})^{-1} \boldsymbol{y}_T \doteq 2\log P(\boldsymbol{y}_T)$, where "$\doteq$" means that we drop determinant terms, thus concentrate on quadratic functions. Since $\log P(\boldsymbol{y}_T) = \sum_t \log P(y_t|\boldsymbol{y}_{<t}) = \sum_t \log N(y_t|\mu_{t-1}(\boldsymbol{x}_t), \sigma_{t-1}^2(\boldsymbol{x}_t) + \sigma^2)$, we have that

$$-\boldsymbol{y}_T^T(\boldsymbol{K}_T + \sigma^2 \boldsymbol{I})^{-1} \boldsymbol{y}_T = -\sum_t \frac{(y_t - \mu_{t-1})^2}{\sigma^2 + \sigma_{t-1}^2}$$
$$= 2\sum_t \varepsilon_t \frac{\mu_{t-1} - f(\boldsymbol{x}_t)}{\sigma^2 + \sigma_{t-1}^2} - \sum_t \frac{\varepsilon_t^2 \sigma_{t-1}^2}{\sigma^2 + \sigma_{t-1}^2} - R$$

with $R = \sum_t (\mu_{t-1} - f(\boldsymbol{x}_t))^2/(\sigma^2 + \sigma_{t-1}^2) \geq 0$. Dropping $-R$ and changing to normalized quantities concludes the proof. ∎

We now define a useful martingale difference sequence. First, it is convenient to define an "escape event" $E_T$ as:

$$E_T = \mathrm{I}\{Z_t \leq \beta_{t+1} \text{ for all } t \leq T\}$$

where $\mathrm{I}\{\cdot\}$ is the indicator function. Define the random variables $M_t$ by

$$M_t = 2\widetilde{\varepsilon}_t E_{t-1} \frac{\widetilde{\mu}_{t-1} - \widetilde{f}(\boldsymbol{x}_t)}{1 + \widetilde{\sigma}_{t-1}^2}.$$

Now, since $\widetilde{\varepsilon}_t$ is a martingale difference sequence with respect to the histories $\mathcal{H}_{<t}$ and $M_t/\widetilde{\varepsilon}_t$ is deterministic given $\mathcal{H}_{<t}$, $M_t$ is a martingale difference sequence as well. Next, we show that with high probability, the associated martingale $\sum_{t=1}^T M_t$ does not grow too large.

**Lemma 7.3** *Given $\delta \in (0, 1)$ and $\beta_t$ as defined in in Theorem 6, we have that*

$$\Pr\left\{\forall T, \ \sum_{t=1}^T M_t \leq \beta_{T+1}/2\right\} \geq 1 - \delta,$$

The proof is given below in Section B.3. Equipped with this lemma, we can prove Theorem 6.

**Proof** [of Theorem 6] It suffices to show that the high-probability event described in Lemma 7.3 is contained in the support of $E_T$ for every $T$. We prove the latter by induction on $T$.

By Lemma 7.2 and the definition of $\beta_1$, we know that $Z_0 \leq \|f\|_k \leq \beta_1$. Hence $E_0 = 1$ always. Now suppose the high-probability event of Lemma 7.3 holds, in particular $\sum_{t=1}^T M_t \leq \beta_{T+1}/2$. For the inductive hypothesis, assume $E_{T-1} = 1$. Using this and Lemma 7.2:

$$Z_T \leq \|f\|_k^2 + 2\sum_{t=1}^T \frac{\widetilde{\varepsilon}_t(\widetilde{\mu}_{t-1} - \widetilde{f}(\boldsymbol{x}_t))}{1 + \widetilde{\sigma}_{t-1}^2} + \sum_{t=1}^T \frac{\widetilde{\varepsilon}_t^2 \widetilde{\sigma}_{t-1}^2}{1 + \widetilde{\sigma}_{t-1}^2}$$
$$= \|f\|_k^2 + \sum_{t=1}^T M_t + \sum_{t=1}^T \widetilde{\varepsilon}_t^2 \frac{\widetilde{\sigma}_{t-1}^2}{1 + \widetilde{\sigma}_{t-1}^2}$$
$$\leq \|f\|_k^2 + \beta_{T+1}/2 + \sum_{t=1}^T \min\{\widetilde{\sigma}_{t-1}^2, 1\}$$
$$\leq \|f\|_k^2 + \beta_{T+1}/2 + (2/\log 2)\gamma_T \leq \beta_{T+1}.$$

The equality in the second step uses the inductive hypothesis. Thus we have shown $E_T = 1$, completing the induction. ∎

### B.3. Concentration

What remains to be shown is Lemma 7.3. While the step sizes $|M_t|$ are uniformly bounded, a standard application of the Hoeffding-Azuma inequality leads to a bound of $T^{3/4}$, too large for our purpose. We use the more specific Theorem 7 instead, which requires to control the conditional variances rather than the marginal variances which can be much larger.

**Proof** [of Lemma 7.3] Let us first obtain upper bounds on the step sizes of our martingale.

$$
\begin{aligned}
|M_t| &= 2|\widetilde{\varepsilon}_t|E_{t-1}\frac{|\widetilde{\mu}_{t-1} - \widetilde{f}(\boldsymbol{x}_t)|}{1+\widetilde{\sigma}_{t-1}^2} \\
&\leq 2|\widetilde{\varepsilon}_t|E_{t-1}\frac{\beta_t^{1/2}\widetilde{\sigma}_{t-1}}{1+\widetilde{\sigma}_{t-1}^2} \\
&\leq 2|\widetilde{\varepsilon}_t|E_{t-1}\beta_t^{1/2}\min\{\widetilde{\sigma}_{t-1}, 1/2\}, \qquad (12)
\end{aligned}
$$

where the first inequality follows from the definition of $E_t$. Moreover, $r/(1+r^2) \leq \min\{r, 1/2\}$ for $r \geq 0$. Therefore, $|M_t| \leq \beta_T^{1/2}$, since $|\widetilde{\varepsilon}_t| \leq 1$ and $\beta_t$ in nondecreasing. Next, we bound the sum of the conditional variances of the martingale:

$$
\begin{aligned}
V_T &:= \sum_{t=1}^T \mathbf{Var}\left(M_t \mid M_1 \dots M_{t-1}\right) \\
&\leq \sum_{t=1}^T 4|\widetilde{\varepsilon}_t|^2 E_{t-1}\beta_t \min\{\widetilde{\sigma}_{t-1}^2, 1/4\} \\
&\leq 4\beta_T \sum_{t=1}^T E_{t-1}\min\{\widetilde{\sigma}_{t-1}^2, 1/4\} \qquad |\widetilde{\varepsilon}_t| \leq 1 \\
&\leq 9\beta_T \gamma_T.
\end{aligned}
$$

In the last line, we used Lemma 7.1 with $\alpha = 1/4$, noting that $8\alpha/\log(1+\alpha) \leq 9$. Since we have established that the sum of conditional variances, $V_T$, is always bounded by $9\beta_T\gamma_T$, we can apply Theorem 7 with parameters $a = \beta_{T+1}/2$, $b = \beta_{T+1}^{1/2}$ and $v = 9\beta_T\gamma_T$ to get

$$
\begin{aligned}
&\Pr\left\{\sum_{t=1}^T M_t \geq \beta_{T+1}/2\right\} \\
&= \Pr\left\{\sum_{t=1}^T M_t \geq \beta_{T+1}/2 \text{ and } V_T \leq 9\beta_T\gamma_T\right\} \\
&\leq \exp\left(\frac{-(\beta_{T+1}/2)^2}{2(9\beta_T\gamma_T) + \frac{2}{3}(\beta_{T+1}/2)\beta_{T+1}^{1/2}}\right) \\
&= \exp\left(\frac{-\beta_{T+1}}{72\gamma_T + \frac{4}{3}\beta_{T+1}^{1/2}}\right) \\
&\leq \max\left\{\exp\left(\frac{-\beta_{T+1}}{144\gamma_T}\right), \exp\left(\frac{-3\beta_{T+1}^{1/2}}{8}\right)\right\}.
\end{aligned}
$$

Note that our choice of $\beta_{T+1}$ satisfies:

$$\max\left\{144\gamma_T\log(T^2/\delta), \left((8/3)\log(T^2/\delta)\right)^2\right\} \leq \beta_{T+1}.$$

Therefore, the previous probability is bounded by $\delta/T^2$, whereas the last inequality follows from the definition of $\beta_{T+1}$. With a final application of the union bound:

$$
\begin{aligned}
&\Pr\left\{\sum_{t=1}^T M_t \geq \beta_{T+1}/2 \text{ for some } T\right\} \\
&\leq \sum_{T \geq 1} \Pr\left\{\sum_{t=1}^T M_t \geq \beta_{T+1}/2\right\} \\
&\leq \sum_{T \geq 2} \delta/T^2 \leq \delta(\pi^2/6 - 1) \leq \delta,
\end{aligned}
$$

completing the proof of Lemma 7.3. ■

## C. Bounds on Information Gain

In this section, we show how to bound $\gamma_T$, the maximum information gain after $T$ rounds, for compact $D \subset \mathbb{R}^d$ (assumptions of Theorem 2) and several commonly used covariance functions. In this section, we assume[4] that $k(\boldsymbol{x}, \boldsymbol{x}) = 1$ for all $\boldsymbol{x} \in D$.

The plan of attack is as follows. First, we note that the argument of $\gamma_T$, $\mathrm{I}(\boldsymbol{y}_A; \boldsymbol{f}_A)$ is a submodular function, so $\gamma_T$ can be bounded by the value obtained by greedy maximization. Next, we use a discretization $D_T \subset D$ with $n_T = |D_T| = T^\tau$ with nearest neighbour distance $o(1)$, consider the kernel matrix $\boldsymbol{K}_{D_T} \in \mathbb{R}^{n_T \times n_T}$, and bound $\gamma_T$ by an expression involving the eigenvalues $\{\hat{\lambda}_t\}$ of this matrix, which is done by a further relaxation of the greedy procedure. Finally, we bound this empirical expression in terms of the kernel operator eigenvalues of $k$ w.r.t. the uniform distribution on $D$. Asymptotic expressions for the latter are reviewed in Seeger et al. (2008), which we plug in to obtain our results. A key step in this argument is to ensure the existence of a discretization $D_T$, for which tails of the empirical spectrum can be bounded by tails of the process spectrum. We will invoke the probabilistic method for that.

### C.1. Greedy Maximization and Discretization

In this section, we fix $T \in \mathbb{N}$ and assume the existence of a discretization $D_T \subset D$, $n_T = |D_T|$ on the order of $T^\tau$, such that:

$$\forall \boldsymbol{x} \in D\ \exists [\boldsymbol{x}]_T \in D_T\ :\ \|\boldsymbol{x} - [\boldsymbol{x}]_T\| = \mathcal{O}(T^{-\tau/d}). \quad (13)$$

We come back to the choice of $D_T$ below. We restrict the information gain to subsets $A \subset D_T$:

$$\widetilde{\gamma}_T = \max_{A \subset D_T, |A|=T} \mathrm{I}(\boldsymbol{y}_A; \boldsymbol{f}_A).$$

Of course, $\widetilde{\gamma}_T \leq \gamma_T$, but we can bound the slack.

---

[4] Without loss in generality. We use this assumption below to ensure that $n_T^{-1}\mathrm{tr}\boldsymbol{K}_{D_T} = \int k(\boldsymbol{x}, \boldsymbol{x})\,d\boldsymbol{x}$. If $k(\boldsymbol{x}, \boldsymbol{x})$ is not constant, this is approximately true by the law of large numbers, and our result below remains valid.

**Lemma 7.4** *Under the assumptions of Theorem 2, the information gain $F_T(\{\boldsymbol{x}_t\}) = (1/2)\log|\boldsymbol{I} + \sigma^{-2}\boldsymbol{K}_{\{\boldsymbol{x}_t\}}|$ is uniformly Lipschitz-continuous in each component $\boldsymbol{x}_t \in D$.*

**Proof** The assumptions of Theorem 2 imply that the kernel $K(\boldsymbol{x}, \boldsymbol{x}')$ is continuously differentiable. The result follows from the fact that $F_T(\{\boldsymbol{x}_t\})$ is continuously differentiable in the kernel matrix $\boldsymbol{K}_{\{\boldsymbol{x}_t\}}$. ∎

**Lemma 7.5** *Let $D_T$ be a discretization of $D$ such that (13) holds. Under the assumptions of Theorem 2, we have that*

$$0 \le \gamma_T - \tilde{\gamma}_T = \mathcal{O}(T^{1-\tau/d}).$$

**Proof** Fix $T \in \mathbb{N}$, and let $A = \{\boldsymbol{x}_1, \ldots, \boldsymbol{x}_T\}$ be a maximizer for $\gamma_T$. Consider neighbours $[\boldsymbol{x}_t]_T \in D_T$ according to (13), $[A]_T = \{[\boldsymbol{x}_t]_T\}$. Then,

$$0 \le \gamma_T - \tilde{\gamma}_T \le \gamma_T - \mathrm{I}(\boldsymbol{y}_{[A]_T}; \boldsymbol{f}_{[A]_T}) = F_T(A) - F_T([A]_T),$$

where $F_T(\{\boldsymbol{x}_t\}) = (1/2)\log|\boldsymbol{I} + \sigma^{-2}\boldsymbol{K}_{\{\boldsymbol{x}_t\}}|$. By Lemma 7.4, $F_T$ is uniformly Lipschitz-continuous in each component, so that $|\gamma_T - \mathrm{I}(\boldsymbol{y}_{[A]_T}; \boldsymbol{f}_{[A]_T})| = \mathcal{O}(T \max_t \|\boldsymbol{x}_t - [\boldsymbol{x}_t]_T\|) = \mathcal{O}(T^{1-\tau/d})$ by (13) and the mean value theorem. ∎

We concentrate on $\tilde{\gamma}_T$ in the sequel. Let $\boldsymbol{K}_{D_T} = [k(\boldsymbol{x}, \boldsymbol{x}')]_{\boldsymbol{x}, \boldsymbol{x}' \in D_T}$ be the kernel matrix over the entire $D_T$, and $\boldsymbol{K}_{D_T} = \boldsymbol{U}\hat{\boldsymbol{\Lambda}}\boldsymbol{U}^T$ its eigendecomposition, with $\hat{\lambda}_1 \ge \hat{\lambda}_2 \ge \cdots \ge 0$ and $\boldsymbol{U} = [\boldsymbol{u}_1\,\boldsymbol{u}_2\,\ldots]$ orthonormal. Here, if $T > n_T$, define $\hat{\lambda}_t = 0$ for $t = n_T + 1, \ldots, T$. Information gain maximization over a finite $D_T$ can be described in terms of a simple linear-Gaussian model over the unknown $\boldsymbol{f} \in \mathbb{R}^{n_T}$, with prior $P(\boldsymbol{f}) = N(\boldsymbol{0}, \boldsymbol{K}_{D_T})$ and likelihood potentials $P(y_t|\boldsymbol{f}) = N(\boldsymbol{v}_t^T \boldsymbol{f}, \sigma^2)$ with unit-norm features, $\|\boldsymbol{v}_t\| = 1$. With the following lemma, we upper-bound $\tilde{\gamma}_T$ by way of two relaxations.

**Lemma 7.6** *For any $T \ge 1$, we have that*

$$\tilde{\gamma}_T \le \frac{1/2}{1-e^{-1}} \max_{m_1, \ldots, m_T} \sum_{t=1}^{T} \log(1 + \sigma^{-2} m_t \hat{\lambda}_t),$$

*subject to $m_t \in \mathbb{N}$, $\sum_t m_T = T$, where $\hat{\lambda}_1 \ge \hat{\lambda}_2 \ge \ldots$ is the spectrum of the kernel matrix $\boldsymbol{K}_{D_T}$. Here, if $T > n_T$, then $m_t = 0$ for $t > n_T$.*

**Proof** As shown by Krause & Guestrin (2005), the function $F(A) = \mathrm{I}(\boldsymbol{y}_A; \boldsymbol{f})$ is submodular. In the particular case considered here, this can be seen as follows: $F(A) = \mathrm{H}(\boldsymbol{y}_A) - \mathrm{H}(\boldsymbol{y}_A \mid \boldsymbol{f})$, where the entropy $\mathrm{H}(\boldsymbol{y}_A)$ is a (not-necessarily monotonic) submodular function in $A$, and since the noise is conditionally independent given $\boldsymbol{f}$, $\mathrm{H}(\boldsymbol{y}_A \mid \boldsymbol{f})$ is an additive (modular) function in $A$. Subtracting a modular function preserves submodularity, thus $F(A)$ is submodular. Furthermore, the information gain is monotonic in $A$ (i.e., $F(A) \le F(B)$ whenever $A \subseteq B$) (Cover & Thomas, 1991). Thus, we can apply the result of Nemhauser et al. (1978)[5] which guarantees that $\tilde{\gamma}_T$ is upper-bounded by $1/(1 - 1/e)$ times the value the greedy maximization algorithm attains. The latter chooses features of the form $\boldsymbol{v}_t = \boldsymbol{\delta}_{\boldsymbol{x}_t} = [\mathrm{I}_{\{\boldsymbol{x} = \boldsymbol{x}_t\}}]$ in each round, $\boldsymbol{x}_t \in D_T$. We upper-bound the greedy maximum once more by relaxing these constraints to $\|\boldsymbol{v}_t\| = 1$ only. In the remainder of the proof, we concentrate on this relaxed greedy procedure. Suppose that up to round $t$, it chose $\boldsymbol{v}_1, \ldots, \boldsymbol{v}_{t-1}$. The posterior $P(\boldsymbol{f}|\boldsymbol{y}_{t-1})$ has inverse covariance matrix $\boldsymbol{\Sigma}_{t-1}^{-1} = \boldsymbol{K}_{D_T}^{-1} + \sigma^{-2}\boldsymbol{V}_{t-1}\boldsymbol{V}_{t-1}^T$, $\boldsymbol{V}_{t-1} = [\boldsymbol{v}_1 \ldots \boldsymbol{v}_{t-1}]$, and the greedy procedure selects $\boldsymbol{v}$ so to maximize the variance $\boldsymbol{v}^T \boldsymbol{\Sigma}_{t-1} \boldsymbol{v}$: the eigenvector corresponding to $\boldsymbol{\Sigma}_{t-1}$'s largest eigenvalue (by the Rayleigh-Ritz theorem). Since $\boldsymbol{\Sigma}_0 = \boldsymbol{K}_{D_T}$, then $\boldsymbol{v}_1 = \boldsymbol{u}_1$. Moreover, if all $\boldsymbol{v}_{t'}$, $t' < t$, have been chosen among $\boldsymbol{U}$'s columns, then by the inverse covariance expression just given, $\boldsymbol{K}_{D_T}$ and $\boldsymbol{\Sigma}_{t-1}$ have the same eigenvectors, so that $\boldsymbol{v}_t$ is a column of $\boldsymbol{U}$ as well. For example, if $\boldsymbol{v}_t = \boldsymbol{u}_j$, then comparing $\boldsymbol{\Sigma}_{t-1}$ and $\boldsymbol{\Sigma}_t$, all eigenvalues other than the $j$-th remain the same, while the latter is shrunk. Therefore, after $T$ rounds of the relaxed greedy procedure: $\boldsymbol{v}_t \in \{\boldsymbol{u}_1, \ldots, \boldsymbol{u}_{\min\{T, n_T\}}\}$, $t = 1, \ldots, T$: at most the leading $T$ eigenvectors of $\boldsymbol{K}_{D_T}$ can have been selected (possibly multiple times). If $m_t$ denotes the number that the $t$-th column of $\boldsymbol{U}$ has been selected, we obtain the theorem statement by a final bounding step. ∎

### C.2. From Empirical to Process Eigenvalues

The final step will be to relate the empirical spectrum $\{\hat{\lambda}_t\}$ to the kernel operator spectrum. Since $\log(1 + \sigma^{-2} m_t \hat{\lambda}_t) \le \sigma^{-2} m_t \hat{\lambda}_t$ in Theorem 7.6, we will mainly be interested in relating the tail sums of the spectra. Let $\mu(\boldsymbol{x}) = \mathcal{V}(D)^{-1} \mathrm{I}_{\{\boldsymbol{x} \in D\}}$ be the uniform distribution on $D$, $\mathcal{V}(D) = \int_{\boldsymbol{x} \in D} d\boldsymbol{x}$, and assume that $k$ is continuous. Note that $\int k(\boldsymbol{x}, \boldsymbol{x}) \mu(\boldsymbol{x}) d\boldsymbol{x} = 1$ by our assumption $k(\boldsymbol{x}, \boldsymbol{x}) = 1$, so that $k$ is Hilbert-

---

[5]While the result of Nemhauser et al. (1978) is stated in terms of finite sets, it extends to infinite sets as long as the greedy selection can be implemented efficiently.

Schmidt on $L_2(\mu)$. Then, Mercer's theorem (Wahba, 1990) states that the corresponding kernel operator has a discrete eigenspectrum $\{(\lambda_s, \phi_s(\cdot))\}$, and

$$k(\boldsymbol{x}, \boldsymbol{x}') = \sum_{s \geq 1} \lambda_s \phi_s(\boldsymbol{x}) \phi_s(\boldsymbol{x}'),$$

where $\lambda_1 \geq \lambda_2 \geq \cdots \geq 0$, and $\mathbb{E}_\mu[\phi_s(\boldsymbol{x})\phi_t(\boldsymbol{x})] = \delta_{s,t}$. Moreover, $\sum_{s \geq 1} \lambda_s^2 < \infty$, and the expansion of $k$ converges absolutely and uniformly on $D \times D$. Note that $\sum_{s \geq 1} \lambda_s = \sum_{s \geq 1} \lambda_s \mathbb{E}_\mu[\phi_s(\boldsymbol{x})^2] = \int K(\boldsymbol{x}, \boldsymbol{x}) \mu(\boldsymbol{x}) \, d\boldsymbol{x} = 1$. In order to proceed from Theorem 7.6, we have to pick a discretization $D_T$ for which (13) holds, and for which $\sum_{t > T_*} \hat{\lambda}_t$ is not much larger than $\sum_{t > T_*} \lambda_t$. With the following lemma, we determine sizes $n_T$ for which such discretizations exist.

**Lemma 7.7** *Fix $T \in \mathbb{N}$, $\delta > 0$ and $\varepsilon > 0$. There exists a discretization $D_T \subset D$ of size*

$$n_T = \mathcal{V}(D)(\varepsilon/\sqrt{d})^{-d}[\log(1/\delta) + d\log(\sqrt{d}/\varepsilon) + \log \mathcal{V}(D)]$$

*which fulfils the following requirements:*

- $\varepsilon$*-denseness: For any $\boldsymbol{x} \in D$, there exists $[\boldsymbol{x}]_T \in D_T$ such that $\|\boldsymbol{x} - [\boldsymbol{x}]_T\| \leq \varepsilon$.*

- *If $\mathrm{spec}(\boldsymbol{K}_{D_T}) = \{\hat{\lambda}_1 \geq \hat{\lambda}_2 \geq \ldots\}$, then for any $T_* = 1, \ldots, n_T$:*

$$n_T^{-1} \sum_{t=1}^{T_*} \hat{\lambda}_t \geq \sum_{t=1}^{T_*} \lambda_t - \delta.$$

**Proof** First, if we draw $n_T$ samples $\tilde{\boldsymbol{x}}_j \sim \mu(\boldsymbol{x})$ independently at random, then $D_T = \{\tilde{\boldsymbol{x}}_j\}$ is $\varepsilon$-dense with probability $\geq 1 - \delta$. Namely, cover $D$ with $N = \mathcal{V}(D)(\varepsilon/\sqrt{d})^{-d}$ hypercubes of sidelength $\varepsilon/\sqrt{d}$, within which the maximum Euclidean distance is $\varepsilon$. The probability of not hitting at least one cell is upper-bounded by $N(1 - 1/N)^{n_T}$. Since $\log(1 - 1/N) \leq -1/N$, this is upper-bounded by $\delta$ if $n_T \geq N \log(N/\delta)$. Now, let $S = n_T^{-1} \sum_{t=1}^{T_*} \hat{\lambda}_t$. Shawe-Taylor et al. (2005) show that $\mathbb{E}[S] \geq \sum_{t=1}^{T_*} \lambda_t$. If $\mathcal{C}$ is the event $\{D_T \text{ is } \varepsilon\text{-dense}\}$, then $\Pr(\mathcal{C}) \geq 1 - \delta$. Since $S \leq n_T^{-1} \mathrm{tr} \boldsymbol{K}_{D_T} = 1$ in any case, we have that $\mathbb{E}[S|\mathcal{C}] \geq \mathbb{E}[S] - \Pr(\mathcal{C}^c) \geq \sum_{t=1}^{T_*} \lambda_t - \delta$. By the probabilistic method, there must exist some $D_T$ for which $\mathcal{C}$ and the latter inequality holds. ∎

The following lemma, the equivalent of Theorem 4 in the context here, is a direct consequence of Lemma 7.6.

**Lemma 7.8** *Let $D_T$ be some discretization of $D$, $n_T = |D_T|$. Then, for any $T_* = 1, \ldots, \min\{T, n_T\}$:*

$$\tilde{\gamma}_T \leq \frac{1/2}{1 - e^{-1}} \max_{r=1,\ldots,T} \Big( T_* \log(r n_T/\sigma^2) + (T - r)\sigma^{-2} \sum_{t=T_*+1}^{n_T} \hat{\lambda}_t \Big).$$

**Proof** We split the right hand side in Lemma 7.6 at $t = T_*$. Let $r = \sum_{t \leq T_*} m_t$. For $t \leq T_*$: $\log(1 + m_t \hat{\lambda}_t/\sigma^2) \leq \log(r n_T/\sigma^2)$, since $\hat{\lambda}_t \leq n_T$. For $t > T_*$: $\log(1 + m_t \hat{\lambda}_t/\sigma^2) \leq m_t \hat{\lambda}_t/\sigma^2 \leq (T-r)\hat{\lambda}_t/\sigma^2$. ∎

The following theorem describes our "recipe" for obtaining bounds on $\gamma_T$ for a particular kernel $k$, given that tail bounds on $B_k(T_*) = \sum_{s > T_*} \lambda_s$ are known.

**Theorem 8** *Suppose that $D \subset \mathbb{R}^d$ is compact, and $k(\boldsymbol{x}, \boldsymbol{x}')$ is a covariance function for which the additional assumption of Theorem 2 holds. Moreover, let $B_k(T_*) = \sum_{s > T_*} \lambda_s$, where $\{\lambda_s\}$ is the operator spectrum of $k$ with respect to the uniform distribution over $D$. Pick $\tau > 0$, and let $n_T = C_4 T^\tau (\log T)$ with $C_4 = 2\mathcal{V}(D)(2\tau + 1)$. Then, the following bound holds true:*

$$\gamma_T \leq \frac{1/2}{1 - e^{-1}} \max_{r=1,\ldots,T} \Big( T_* \log(r n_T/\sigma^2) + C_4 \sigma^{-2}(1 - r/T)(\log T)\left(T^{\tau+1} B_k(T_*) + 1\right) \Big) + \mathcal{O}(T^{1-\tau/d})$$

*for any $T_* \in \{1, \ldots, n_T\}$.*

**Proof** Let $\varepsilon = d^{1/2} T^{-\tau/d}$ and $\delta = T^{-(\tau+1)}$. Lemma 7.7 provides the existence of a discretization $D_T$ of size $n_T$ which is $\varepsilon$-dense, and for which $n_T^{-1} \sum_{t=1}^{T_*} \hat{\lambda}_t \geq \sum_{t=1}^{T_*} \lambda_t - \delta$. Since $n_T^{-1} \sum_{t=1}^{n_T} \hat{\lambda}_t = 1 = \sum_{t \geq 1} \lambda_t$, then $\sum_{t > T_*} \hat{\lambda}_t \leq B_k(T_*) + \delta$. The statement follows by using Lemma 7.8 with these bounds, and finally employing Lemma 7.5. ∎

### C.3. Proof of Theorem 5

In this section, we instantiate Theorem 8 in order to obtain bounds on $\gamma_T$ for Squared Exponential and Matérn kernels, results which are summarized in Theorem 5.

SQUARED EXPONENTIAL KERNEL

For the Squared Exponential kernel $k$, $B_k(T_*)$ is given by Seeger et al. (2008). While $\mu(\boldsymbol{x})$ was Gaussian

there, the same decay rate holds for $\lambda_s$ w.r.t. uniform $\mu(\bm{x})$, while constants might change. In hindsight, it turns out that $\tau = d$ is the optimal choice for the discretization size, rendering the second term in Theorem 5 to be $\mathcal{O}(1)$, which is subdominant and will be neglected in the sequel. We have that $\lambda_s \leq cB^{s^{1/d}}$ with $B < 1$. Following their analysis,

$$B_k(T_*) \leq c(d!)\alpha^{-d}e^{-\beta}\sum_{j=0}^{d-1}(j!)^{-1}\beta^j,$$

where $\alpha = -\log B$, $\beta = \alpha T_*^{1/d}$. Therefore, $B_k(T_*) = \mathcal{O}(e^{-\beta}\beta^{d-1})$, $\beta = \alpha T_*^{1/d}$.

We have to pick $T_*$ such that $e^{-\beta}$ is not much larger than $(Tn_T)^{-1}$. Suppose that $T_* = [\log(Tn_T)/\alpha]^d$, so that $e^{-\beta} = (Tn_T)^{-1}$, $\beta = \log(Tn_T)$. The bound becomes

$$\max_{r=1,\ldots,T}\Big(T_*\log(rn_T/\sigma^2) + \sigma^{-2}(1-r/T)(C_5\beta^{d-1} + C_4(\log T))\Big)$$

with $n_T = C_4 T^d(\log T)$. The first part dominates, so that $r = T$ and $\gamma_T = \mathcal{O}([\log(T^{d+1}(\log T))]^{d+1}) = \mathcal{O}((\log T)^{d+1})$. This should be compared with $\mathbb{E}[\mathrm{I}(\bm{y}_T;\bm{f}_T)] = \mathcal{O}((\log T)^{d+1})$ given by Seeger et al. (2008), where the $\bm{x}_t$ are drawn independently from a Gaussian base distribution. At least restricted to a compact set $D$, we obtain the same expression to leading order for $\max_{\{\bm{x}_t\}} \mathrm{I}(\bm{y}_T;\bm{f}_T)$.

Matérn Kernels

For Matérn kernels $k$ with roughness parameter $\nu$, $B_k(T_*)$ is given by Seeger et al. (2008) for the uniform base distribution $\mu(\bm{x})$ on $D$. Namely, $\lambda_s \leq cs^{-(2\nu+d)/d}$ for almost all $s \in \mathbb{N}$, and $B_k(T_*) = \mathcal{O}(T_*^{1-(2\nu+d)/d})$. To match terms in the $\tilde{\gamma}_T$ bound, we choose $T_* = (Tn_T)^{d/(2\nu+d)}(\log(Tn_T))^\kappa$ ($\kappa$ chosen below), so that the bound becomes

$$\max_{r=1,\ldots,T}\Big(T_*\log(rn_T/\sigma^2) + \sigma^{-2}(1-r/T)$$
$$\times (C_5 T_*(\log(Tn_T))^{-\kappa(2\nu+d)/d} + C_4(\log T))\Big)$$
$$+ \mathcal{O}(T^{1-\tau/d})$$

with $n_T = C_4 T^\tau(\log T)$. For $\kappa = -d/(2\nu+d)$, we obtain that the maximum over $r$ is $\mathcal{O}(T_*\log(Tn_T)) = \mathcal{O}(T^{(\tau+1)d/(2\nu+d)}(\log T))$. Finally, we choose $\tau = 2\nu d/(2\nu+d(d+1))$ to match this term with $\mathcal{O}(T^{1-\tau/d})$. Plugging this in, we have $\gamma_T = \mathcal{O}(T^{1-2\eta}(\log T))$, $\eta = \frac{\nu}{2\nu+d(d+1)}$. Together with Theorem 2 (for $\nu > 2$), we have that $R_T = \mathcal{O}^*(T^{1-\eta})$ (suppressing log factors): for any $\nu > 2$ and any dimension $d$, the GP-UCB algorithm is guaranteed to be no-regret in this case with arbitrarily high probability.

How does this bound compare to the bound on $\mathbb{E}[\mathrm{I}(\bm{y}_T;\bm{f}_T)]$ given by Seeger et al. (2008)? Here, $\gamma_T = \mathcal{O}(T^{d(d+1)/(2\nu+d(d+1))}(\log T))$, while $\mathbb{E}[\mathrm{I}(\bm{y}_T;\bm{f}_T)] = \mathcal{O}(T^{d/(2\nu+d)}(\log T)^{2\nu/(2\nu+d)})$.

Linear Kernel

For linear kernels $k(\bm{x},\bm{x}') = \bm{x}^T\bm{x}'$, $\bm{x} \in \mathbb{R}^d$ with $\|\bm{x}\| \leq 1$, we can bound $\gamma_T$ directly. Let $\bm{X}_T = [\bm{x}_1 \ldots, \bm{x}_T] \in \mathbb{R}^{d\times T}$ with all $\|\bm{x}_t\| \leq 1$. Now,

$$\log|\bm{I} + \sigma^{-2}\bm{X}_T^T\bm{X}_T| = \log|\bm{I} + \sigma^{-2}\bm{X}_T\bm{X}_T^T|$$
$$\leq \log|\bm{I} + \sigma^{-2}\bm{D}|$$

with $\bm{D} = \operatorname{diag}\operatorname{diag}^{-1}(\bm{X}_T\bm{X}_T^T)$, by Hadamard's inequality. The largest eigenvalue $\hat{\lambda}_1$ of $\bm{X}_T\bm{X}_T^T$ is $\mathcal{O}(T)$, so that

$$\log|\bm{I} + \sigma^{-2}\bm{X}_T^T\bm{X}_T| \leq d\log(1+\sigma^{-2}\hat{\lambda}_1),$$

and $\gamma_T = \mathcal{O}(d\log T)$.